# An Autonomous Reactive Architecture for Efficient AUV Mission Time Management in Realistic Severe Ocean Environment

Somaiyeh Mahmoud.Zadeh, David M.W Powers, Karl Sammut

Centre for Maritime Engineering, Control and Imaging,
School of Computer Science, Engineering and Mathematics, Flinders University, Adelaide, SA 5042, Australia
somaiyeh.mahmoudzadeh@flinders.edu.au
david.powers@flinders.edu.au
karl.sammut@flinders.edu.au

**Abstract** Today AUVs operation still remains restricted to very particular tasks with low real autonomy due to battery restrictions. Efficient motion planning and mission scheduling are principle requirement toward advance autonomy and facilitate the vehicle to handle long-range operations. A single vehicle cannot carry out all tasks in a large scale terrain; hence, it needs a certain degree of autonomy in performing robust decision making and awareness of the mission/environment to trade-off between tasks to be completed, managing the available time, and ensuring safe deployment at all stages of the mission. In this respect, this research introduces a modular control architecture including higher/lower level planners, in which the higher level module is responsible for increasing mission productivity by assigning prioritized tasks while guiding the vehicle toward its final destination in a terrain covered by several waypoints; and the lower level is responsible for vehicle's safe deployment in a smaller scale encountering time-varying ocean current and different uncertain static/moving obstacles similar to actual ocean environment. Synchronization between higher and lower level modules is efficiently configured to manage the mission time and to guarantee on-time termination of the mission. The performance and accuracy of two higher and lower level modules are tested and validated using ant colony and firefly optimization algorithm, respectively. After all, the overall performance of the architecture is investigated in 10 different mission scenarios. The analyze of the captured results from different simulated missions confirm the efficiency and inherent robustness of the introduced architecture in efficient time management, safe deployment, and providing beneficial operation by proper prioritizing the tasks in accordance with mission time.

***Keywords*** *Autonomous vehicle, Hierarchal architecture, Mission/motion/path planning, Underwater missions, Task assignment, Time management, Decision Autonomy*

| **Nomenclature** | | | |
|---|---|---|---|
| $P^i_{x,y,z}$ | Position of arbitrary waypoint *i* in 3-D space | $\{b\}$ | AUVs Body-Fixed frame |
| $Task_{ij}$ | Task index corresponding to edge $q_{ij}$ | $\eta$ | The AUV state in the $\{n\}$ frame |
| $d_{ij}$ | Distance between location of $P^i_{x,y,z}$ and $P^j_{x,y,z}$ | $\upsilon$ | Vehicle's velocity vector in the $\{b\}$ frame |
| $t_{ij}$ | Estimated time for passing the $d_{ij}$ | $u,v,w$ | Linear components of $\upsilon$: surge, sway, heave |
| $w_{ij}$ | The weight of assigned $Task_{ij}$ to $q_{ij}$ | $X,Y,Z$ | Position of the AUV along the path $\wp$ in $\{n\}$ frame |
| $T_M$ | Estimated mission time | $\phi,\theta,\psi$ | AUV's Euler angles of roll, pitch, yaw in $\{n\}$ frame |
| $T_{Total}$ | The total available time based on battery capacity | $p,q,r$ | AUV's angular velocities in the *x-y-z* axis |
| $\Theta_p$ | Position of object $\Theta$ | $\wp$ | The potential path produced by LOP-P |
| $\Theta_r$ | Radius of object $\Theta$ | $B_{i,K}$ | Blending functions used to generate B-Spline curve |
| $\Theta_{UR}$ | Object's uncertainty rate | $\vartheta$ | Control point for B-Spline curve |
| $V_C$ | Water current velocity | $L_\wp$ | The path length |
| $u_c,v_c,w_c$ | *x,y,z* components of water current vector | $T_\wp$ | The path time |
| $\Gamma_{3\text{-}D}$ | Symbol of the 3-D volume | $T_\varepsilon$ | Time expected for traversing an edge $\approx t_{ij}$ |
| $S$ | Symbol of the 2-D x-y space | $\wp_{Cost}$ | The cost of local path generated by LOP-P |
| $S^o$ | Center of the current vortex | $M_{Cost}$ | The mission cost including $C_\wp$ |
| $\ell$ | Radius of the current vortex | $V_{\sum M,\Theta}$ | Collision violation |
| $\Im$ | Strength of the current vortex | $V_\wp$ | kinodynamic constraints Violation |
| $\lambda_w$ | Covariance matrix for radius of vortex | $\wp_{CPU}$ | computational time for generating a local path |
| $U_R^C(t)$ | Current update rate at time *t* | $M_{CPU}$ | The mission planning CPU time |
| $\{n\}$ | North-East-Depth (NED) frame | $T_{compute}$ | Computation time for checking mission re-planning criterion |

## 1 Introduction

Most of current AUVs applications are supervised from the support vessel that provides higher-level decisions in critical situation and generally takes enormous cost during the mission [1]. A growing attention has been devoted in recent years on increasing the ranges of missions, vehicles endurance, extending vehicles applicability, promoting vehicles autonomy to handle longer missions without supervision, and reducing operation costs [2]. AUVs should operate in turbulent undersea environment with complex spatio-temporal variability, where the current variability can perturb AUVs safety conditions. Currents instabilities can influence vehicle's motion and probably push it to an undesired direction. Robustness of AUV to this strong environmental variability is a key element to mission performance and carry out safety considerations. Restrictions of priori knowledge about later conditions of the environment reduce AUVs autonomy and robustness. Additionally, AUVs capabilities in handling mission objectives is directly influenced by performance of task allocation system. Many of the today's AUV missions is limited to executing a list of pre-programmed instructions and completing a predefined sequences of tasks. Autonomous adaptation of AUVs in performing different tasks in dynamic and continuously changing environment has not been completely fulfilled yet and it is still necessary for the operators to remain in the loop of considering and making decisions [3].

Generally, the AUVs control architecture consists of two different execution layers based on its control requirements i.e. deliberative and reactive layers [4]. The deliberative layer manages the concurrent execution of several tasks with different priorities. The reactive layer manages real-time reactions to perform quick respond to critical events. These automatic functions are executed in the background and promote AUVs self-management characteristics. Mission timing and AUV's time management is a fundamental requirement toward mission success. Consequently, reaching higher performance in real-time applications is a challenging issue due to limited resource availability and the dynamic changes of the environment. Thus, the vehicle should trade-off within the problem constraints and mission productivity while managing the available time and risks. Evidently, decision autonomy and vehicles routing/task assignment capability are connected in various standpoints. On the other hand, instantly adapting to the continuously changing situations of the real world where the most of the information are uncertain and unknown is strongly critical. Accordingly, in lower level also, the vehicle should be capable of copying dynamic time variant ocean current and avoiding collision to different static and dynamic obstacles to ensure safe deployment during the mission. To satisfy the addressed requirements and handling these challenges, development of more evolved embedded functions are required, which can promote the platform's autonomy in both higher and lower levels while maintaining the trust on vehicles safety at all stages of the mission. Many efforts have been devoted in recent years for promoting AUVs' autonomy in mission task assignment in recent years; however, there are still many challenges on achieving a satisfactory level of autonomy for AUV in this regard. Very few investigations are funded to address issues with both low and high level autonomy development and the problem remains unsolved for underwater missions. According to these fundamental concepts, various methodologies have been discussed and worked out in the literature as follows.

### 1.1 Previous Works on Vehicle Routing Problem (VRP) and Task Assignment Approach

Effective routing has a great impact on vehicles time management as well as mission performance by appropriate selection and arrangement of the tasks sequence. Liu and Bucknall (2015) proposed a three-layer structure to facilitate multiple unmanned surface vehicles to accomplish task management and formation route planning in a maritime environment [5]. Eichhorn (2015) implemented graph-based methods for the AUV ''SLOCUM Glider'' motion planning in a dynamic environment [6]. The author employed modified Dijkstra Algorithm where the applied modification and conducted time variant cost function simplifies the determination of a time-optimal route in the geometrical graph. An energy efficient fuzzy based route planning system is presented by Kladis et al., (2011) for UAVs motion planning in a graph-like terrain using priori known wind information [7]. Other methods also studied on efficient task assignment for single/multiple vehicle task assign/routing problem such as graph matching algorithm [8], partitioning method [9], Tabu search algorithm [10], branch and cut algorithm [11], A* search algorithm [12-14], Dijkstra [6,15], and evolutionary algorithms [8,16,17]. The traditional algorithms used for graph routing problem such as Dijkstra, or dynamic programming algorithm has major shortcomings such as high computational complexity for real-time applications. Another issue is that all outline literature in this scope mostly focus on routing problem for different unmanned vehicles in which task allocation is the principle direction of these papers and quality of deployment (motion) has not been addressed. The vehicle's safe and confident deployment is a critical factor affecting vehicle's level of autonomy and should be taken into consideration at all stages of the mission in a vast and uncertain environment. So, in the second category, existing AUVs' trajectory/path planning approaches have been discussed, which are more concentrated on quality of vehicle's motion encountering dynamicity of the terrain.

### 1.2 Path/Trajectory Planning Approaches

Path planning itself is a complicated multi-objective Non-deterministic Polynomial-time (NP) hard problem that has a great impact on vehicle's overall autonomy. The path planning strategies proposed in most of the previous researches particularly deal with vehicle's guidance toward the destination encountering dynamic changes of the terrain. Different methods like D*, A*, FM, FM* algorithms have been employed for AUV optimum path generation [18-21]. Particularly, the main drawback of all above-mentioned methods is that their time complexity increases exponentially with increasing the problem space, which is inappropriate for real-time applications. Evolutionary algorithms are population based optimization methods applied successfully on path planning problem that are advantaged to be implemented on a parallel machine with multiple processors, which speeds up the computation process [22,23]. The Particle Swarm Optimization (PSO) [22, 24], Evolution (DE) [24, 25], Genetic Algorithm (GA) [22, 26, 27], multi-objective genetic algorithm (NSGA II) [28], Differential], and Quantum-based PSO (QPSO) [29] are some popular types of optimization algorithms applied successfully on offline/online path planning approaches. Although various path planning techniques have been suggested for autonomous vehicles, AUV-oriented applications still face several difficulties. Path replanning in this context requires a significant computational effort for progressing data from update of entire terrain, which is problematic in large scale operations. Accurate prediction of the behavior of a dynamic large-scale terrain, far-off the vehicles sensor coverage is unreliable and impractical as only awareness of environment in vicinity of the vehicle is sufficient such that the vehicle can be able to perform reaction to prompt changes. Another problem is that the path planning strategies are not designed for handling the task assignment and time management in a graph-like terrain with respect to graph routing restrictions in cases that the vehicle is required to carry out a specific sequence of ordered tasks. In these kind of problems a routing strategy is a proper approach in dealing with graph search constraints for the VRP problem and organizing the task. However, routing strategies are not accurate enough in copping dynamic variations of the terrain, but these approaches fracture the operation area to smaller beneficent sections and give overview of the pathway for vehicle's maneuver that helps reducing the computational load. Accordingly, each approach is able to handle only a specific level of autonomy in terms of organizing the tasks and managing the mission or dealing with variations of the environment and producing a high quality maneuver for the vehicle. This research addresses the AUV's requirements in both high and low level autonomy toward making AUV systems more intelligent and robust in managing its availabilities and adaption to a dynamic environment.

### 1.3 Research Assumptions and Contributions

A novel autonomous architecture with online re-planning capability is developed to carry out the underwater missions in a large scale environment in the presence of severe environmental disturbances. The system incorporates two different execution layers, deliberative and reactive, to satisfy the AUV's high and low level autonomy requirements in motion planning and mission management. A "Task Organize Mission Plan" (TOM-P) module is designed for the deliberative layer to provide a higher level of autonomy by ordering the execution of several tasks so that the AUV gets directed toward the targeted location. The TOM-P module in the top level simultaneously rearranges the order of tasks taking passage of time into account. The reactive layer, on the other hand, is responsible for performing real-time reactions and quick response to critical events. A "Local On-line Path-Plan" (LOP-P) module was developed in this layer and includes automatic functions that executed in the background to promote the AUV's self-management characteristics. The LOP-P module at the lower level deals with quality and safety of deployment along the ordered tasks (produced by higher layer TOM-P) where persistent variation of the sub-area in proximity of the vehicle is considered simultaneously. The operation of each module may take longer than expectation due to unexpected dynamic changes in the environment. In order to reclaim the missed time an efficacious synchronization scheme (named "Synchron") is added to the architecture to keep pace of modules in different layers of the system.

This research is a completion of the previous attempts [30-32] that construction of the mission-motion planners [31,32] is boosted to a consistent structure in scheme of a coherent architecture in which the performance of the proposed construction is mainly independent of the employed algorithms by modules. The main reason for remarkable performance of this system is the fashion of mixing and matching two disparate strategy from two different prospective and constructing an accurate synchronization between them. A significant benefit of such modular construction is that the modules can employ different methods or their functionality get upgraded without manipulating the system's structure. This advantage specifically increases the reusability and versatility the control architecture and eases updating/upgrading AUV's functionalities to be compatible with other applications, and in particular implement it for other autonomous systems such as unmanned aerial, ground or surface vehicles.

The previous study [30], which provided a primary basic idea of such a modular framework, is expanded by enhancing the synchronization process and maneuverability of the vehicle by upgrading the path planner with the added capability of reactive re-planning, generalizing the applicability of the planners by modeling more realistic underwater situations incorporating kinodynamics of the AUV, variable water currents and uncertainty of the terrain.

The paper is prepared in following sections: the mechanism of the TOM-P and LOP-P modules are demonstrated in Section 2, respectively. An overview of the adopted methods by both of modules on corresponding problems is presented in Section 3. The architectures evaluation and analyze of the simulation results is provided in Section 4 and the conclusion of this research is provided by Section 5.

## 2 Proposed Modular Control Architecture

Proposed architecture designed in separate modules running concurrently including TOM-P in top level with higher level of decision autonomy, and the LOP-P in lower level to autonomously carry out the collision avoidance, cope with current force and handle similar environmental challenges. The modules interact simultaneously by back feeding the situational awareness of the surrounding operating field; accordingly the system decides to re-plan the path or mission or continue the current mission. Hence, the Synchron module is designed to manage the lost time within the LOP-P/ TOM-P process and reactively adapt the system to the last update of environment and decision parameters (e.g. remaining time). This process frequently repeated until the AUV reaches to the end point. Real map data and different uncertain dynamic/ static objects are encountered to model a realistic marine environment.

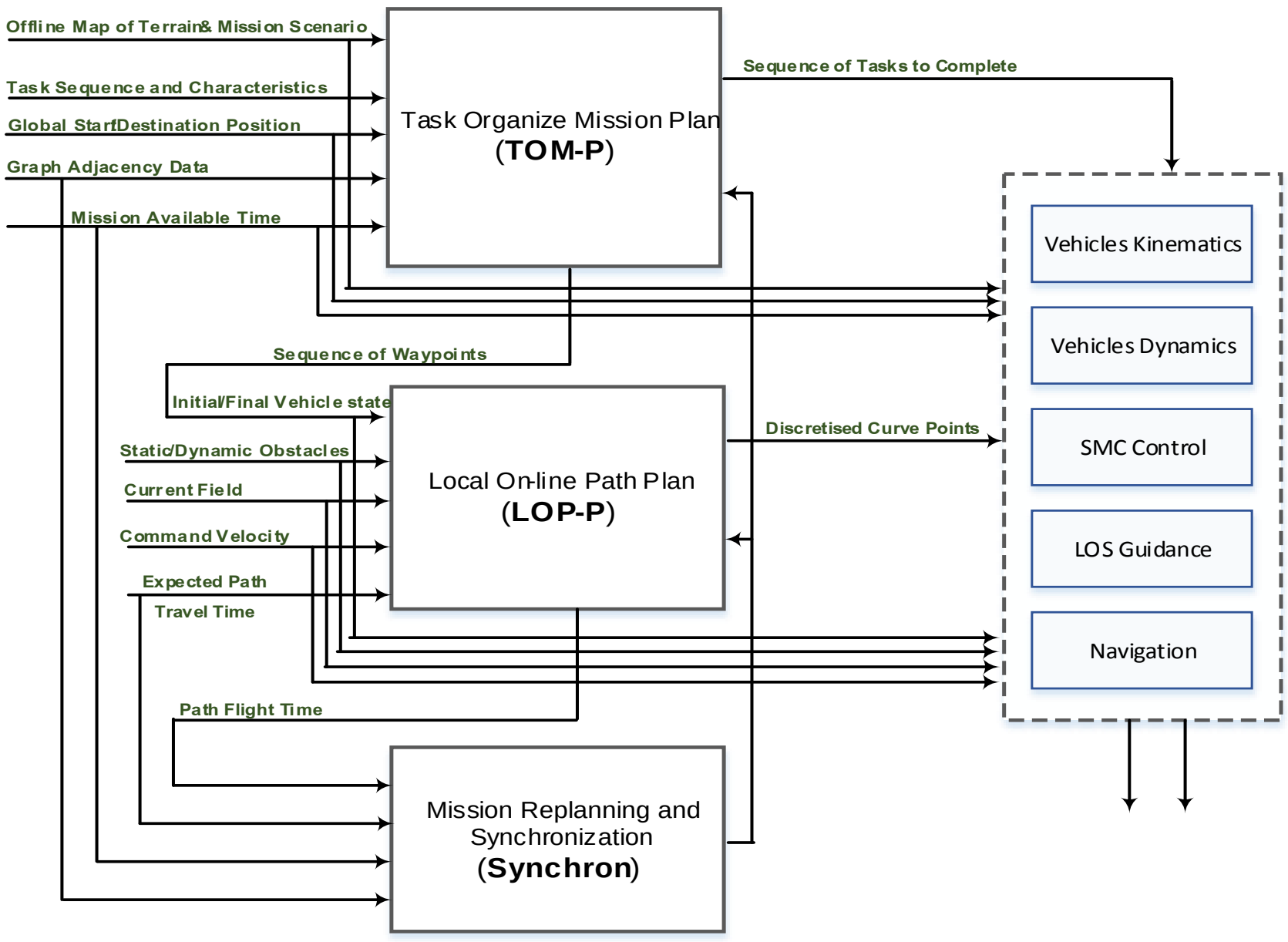

**Fig.1.** The information flow in proposed control architecture including TOM-P, LOP-P, and Synchron modules

Besides, impact of the time variant water current on the AUV's motion is incorporated. To cope dynamic variations of the environment, the LOP-P is facilitated to perform on-line re-planning from vehicles location to any predefined target point. The conceptual and mathematical model of the modular framework including TOM-P, LOP-P and Synchron modules is presented in following subsections. The schematic representation of propounded control architecture is provided by *Fig*.1.

## 2.1 Task Organize Mission Plan (TOM-P)

The terrain modelled as waypoint cluttered operation network, in which various tasks are sparsely distributed on edges of the network and their priority presented by a weight value that initialized once in advance before starting the mission. An example of such an operation network in given by *Fig*.2. In such a network, the edges that are not assigned with a task get value of one, otherwise they get value greater than one proportional to the weight of the corresponding task. Placement of the waypoints in the network performed with a uniform distribution in joint water covered sections of the map and distributed once in advance. The AUV is requested to complete its mission furnishing maximum number of tasks with higher weight while moving toward the destination. Obviously, a single AUV is not able to meet all specified tasks in one mission due to its time and energy restrictions, specifically when operating in a large-scale operation area. Hence, organizing and arranging the tasks involves a decision making procedure, in which a tradeoff exists between ordering the tasks under specified constraints and guiding the vehicle through the waypoints. Therefore, tasks should be arranged in a way that selected edges (tasks) govern the AUV to the destination and meet the time restriction, which is a joint discrete and syndetic multi-objective problem resembling both Knapsack problem and TSP. In this context, the TOM-P module at top level simultaneously tends to determine the optimum order of edges that mathematically described as follows:

$$\forall \quad P^i_{x,y} \sim \mathbf{U}(0{,}10000); \quad P^i_z \sim \mathbf{U}(0{,}100); \quad P^i_{x,y,z} \notin \{Map = 0\}$$

$$\forall \quad P^j_{x,y,z}, P^i_{x,y,z}, \quad \exists q_{ij} : (d_{ij}, t_{ij}, w_{ij}) \Rightarrow \quad q_{ij} : P^i_{x,y,z} \rightarrow P^j_{x,y,z}$$

$$q_{ij} : \begin{cases} d_{ij} = \sqrt{\left(P^j_x - P^i_x\right)^2 + \left(P^j_y - P^i_y\right)^2 + \left(P^j_z - P^i_z\right)^2} \\ t_{ij} = d_{ij} / |\upsilon| \end{cases} \qquad (1)$$

$$\forall \quad q_{ij}, \quad \exists \quad Task_{ij} \equiv w_{ij}$$

$$T_M = \sum_{\substack{i=0 \\ j \neq i}}^{n} lq_{ij} \times t_{ij} = \sum_{\substack{i=0 \\ j \neq i}}^{n} \frac{lq_{ij} \times d_{ij}}{|\upsilon|}, \quad l \in \{0,1\} \qquad (2)$$

$$\max\left( \sum_{\substack{i=0 \\ j \neq i}}^{n} lq_{ij} \times w_{ij} \right) \quad \& \quad \begin{array}{l} \min(|T_M - T_{Total}|) \\ s.t. \\ \max(T_M) < T_{Total} \end{array} \qquad (3)$$

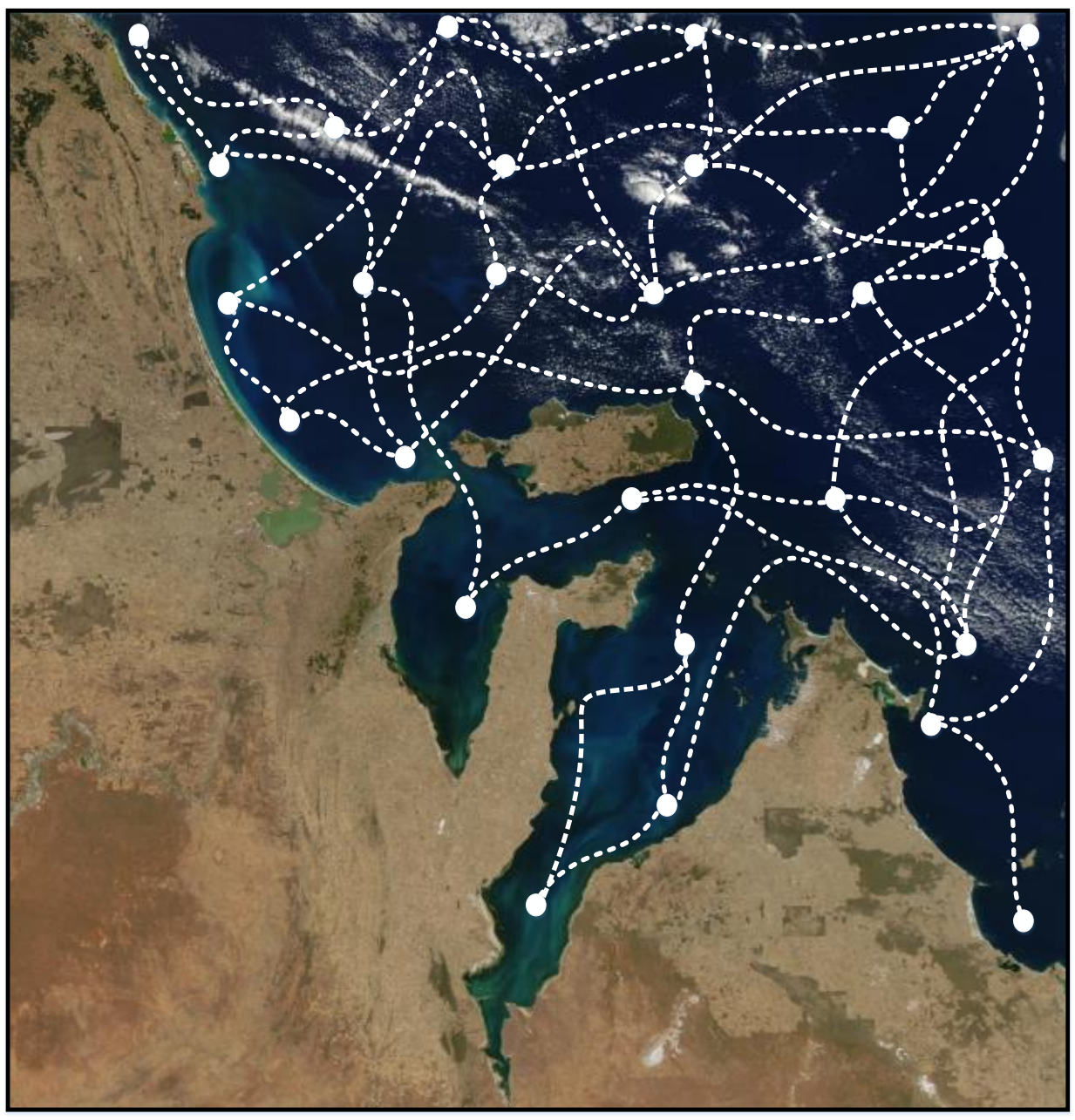

**Fig.2.** A graph representation of operating area covered by waypoints

Definition of all parameters are given by "**Nomenclature**" table. Coordinates of waypoints in the geographical frame shouldn't have any intersection with coastal area denoted by $\{Map=0\}$. The $l$ is the selection variable that shows the selected edges in the network. The total captured weight in a mission should be maximized and the mission time should approach the total available time. Safe deployment is another important task that should be taken into consideration at all stages, which is extremely critical and complicated issue in a vast and uncertain environment that is investigated in subsection 2.2.

## 2.2 Local On-line Path Plan (LOP-P)

### 2.2.1 Modelling the Ocean Terrain, Obstacles and Dynamic Current Field

The ocean environment modelled as a three dimensional volume $\Gamma_{3D}$ covered by uncertain, static/dynamic objects and fixed waypoints. An underwater mission is commenced at a specified starting point and it is terminated when the AUV reaches to a predefined destination point (dock for example). A real map data of Adelaide-gulf st Vincent (the southern coast of Australia) [33] has been used to make the simulation process more realistic. Additional to offline map, two types of static uncertain and moving uncertain obstacles are considered in this research to cover different possibilities of the real world situations. The sonar sensors measure obstacle's coordinates and velocity with a specific uncertainty that modelled with a Gaussian distribution. The obstacle is shown by three components of position ($\Theta_p$), radius ($\Theta_r$) and uncertainty rate ($\Theta_{UR}$). The static uncertain objects modelled with uncertain radius that is varied in a predefined bound with a normal distribution $\Theta_r \sim (\Theta_p, \sigma_0)$. Uncertain moving obstacles also modelled in this research, in which the object motion is affected by current flow. The effect of current presented by uncertainty propagation proportional to current magnitude $\Theta_{UR} \propto |V_C|$ in a circular format radiating out from the center of the object, denoted by (5). The objects position in both cases is initialized using normal distribution bounded to position of start (a) and target (b) waypoints; hence, its probability density function is defined by (4) as follows:

$$\Theta_p^i \in \left[P_{x,y,z}^a, P_{x,y,z}^b\right] - \Theta_r^i$$
$$f(\Theta_p^i;0,\Theta_r^i,P_{x,y,z}^a,P_{x,y,z}^b) = \frac{\Theta_p^i}{\left(\Theta_r^i\right)^2} \Bigg/ \left(\frac{P_{x,y,z}^b - P_{x,y,z}^a}{\Theta_r^i}\right) \tag{4}$$

$$\Theta_p(t) = \Theta_p(t-1) \pm \mathbf{U}(\Theta_{p0},\sigma)$$
$$\Theta_r(t) = B_1\Theta_r(t-1) + B_2 X_{(t-1)} + B_3 \tag{5}$$
$$B_1 = \begin{bmatrix} 1 & \Theta_{UR}(t) & 0 \\ 0 & 1 & 0 \\ 0 & 0 & 1 \end{bmatrix}, B_2 = \begin{bmatrix} 0 \\ 1 \\ 1 \end{bmatrix}, B_3 = \begin{bmatrix} 0 \\ 0 \\ \Theta_{UR}(t) \end{bmatrix}$$

where $X_{(t-1)} \sim \mathcal{N}(0, \sigma_0)$ is the normal distribution that assigned to each obstacle and gets updated iteratively.

On the other hand, the ocean current information can be obtained from remote devices (e.g. satellite observations) or numerical estimation models. Different predictive ocean current models have been constructed previously by [34,35]. A 3D turbulent time varying current model is constructed by this research, where the current update is modelled by multiple layered 2D current structure, in which the current circulation pattern gradually changes in each layer, given by (6):

$$V_c : (u_c, v_c, w_c) = f(\vec{S}^O, \Im, \ell)$$
$$u_c(\vec{S}) = -\Im \frac{y - y_0}{2\pi(\vec{S} - \vec{S}^O)^2}\left[1 - e^{\frac{-(\vec{S}-\vec{S}^O)^2}{\ell^2}}\right]$$
$$v_c(\vec{S}) = \Im \frac{x - x_0}{2\pi(\vec{S} - \vec{S}^O)^2}\left[1 - e^{\frac{-(\vec{S}-\vec{S}^O)^2}{\ell^2}}\right] \tag{6}$$
$$w_c(\vec{S}) = \gamma \Im \frac{1}{\sqrt{\det(2\pi \lambda_w)}} \times e^{\frac{-(\vec{S}-\vec{S}^O)^T}{2\lambda_w(\vec{S}-\vec{S}^O)}}$$
$$\lambda_w = \begin{bmatrix} \ell & 0 \\ 0 & \ell \end{bmatrix}; \quad X^{wc} \sim N(S^o, \lambda_w)$$

A parameter $\gamma$ is used to scale the vertical profile of the current ($w_c$) from the horizontal profile ($u_c$,$v_c$) due to weak vertical motions of the ocean current. To estimate continuous circulation patterns of each subsequent layer, a recursive application of Gaussian noise is applied to the $S^o$,$\Im$, and $\ell$ parameters of 2D case as follows:

$$S_t^o = A_1 S_{t-1}^o + A_2 X_{(t-1)}^{S_x} + A_3 X_{(t-1)}^{S_y}$$
$$\ell(t) = A_1 \ell(t-1) + A_2 X_{(t-1)}^{\ell}$$
$$\Im(t) = A_1 \Im(t-1) + A_2 X_{(t-1)}^{\Im} \tag{7}$$
$$A_1 = \begin{bmatrix} 1 & 0 \\ 0 & 1 \end{bmatrix}, A_2 = \begin{bmatrix} U_R^C(t) \\ 0 \end{bmatrix}, A_3 = \begin{bmatrix} 0 \\ U_R^C(t) \end{bmatrix}$$

where $X^{Sx}$, $X^{Sy}$, $X^{\ell}$, $X^{\Im}$ are Gaussian normal distributions.

### 2.2.2 Modelling Kinematics and Dynamics of the AUV

To describe vehicles motion with six degrees of freedom (6-DOF) two coordinate systems of {$b$} frame, which is located at the centre of buoyancy of the vehicle, and "Earth-Fixed coordinate" of {$n$} frame are utilized [36]. It is assumed that the centres of both coordinates are congruous and the Earth's rotation is ignored. The state variables of {$b$}-{$n$} frames and AUV's kinematics are described by (8) and (9), respectively:

$$\begin{aligned} \{b\} &\mapsto \upsilon : (u, v, w, p, q, r) \\ \{n\} &\mapsto \eta : (X, Y, Z, \varphi, \theta, \psi) \end{aligned} \tag{8}$$

$$\begin{bmatrix} \dot{X} \\ \dot{Y} \\ \dot{Z} \end{bmatrix} = \begin{bmatrix} \cos\psi\cos\theta & -\sin\psi & \cos\psi\sin\theta \\ \sin\psi\cos\theta & \cos\psi & \sin\psi\sin\theta \\ -\sin\theta & 0 & \cos\theta \end{bmatrix} \begin{bmatrix} u \\ v \\ w \end{bmatrix} \tag{9}$$

### 2.2.3 Mathematical Modelling of the Path Planning Problem and Path Re-planning

The path planning is a constraint optimization problem in which the main goal is to provide a safest and quickest trajectory between two points. The study takes the use of B-Spline function to generates the potential trajectories $\wp_i$. The B-Spline curves captured from a set of control points like $\vartheta \subseteq \Gamma_{3\text{-}D}$. Placement of these control points play a substantial role in determining the optimal path. All $\vartheta$ should be located in respective search space limited to predefined Upper-Lower bounds of $\vartheta \in [U_\vartheta, L_\vartheta]$ in Cartesian coordinates. Accordingly, path curve is generated as follows:

$$L_\vartheta \equiv P_{x,y,z}^a \quad \& \quad U_\vartheta \equiv P_{x,y,z}^b$$
$$\begin{cases} \vartheta_{x(i)} = L_{\vartheta(x)}^i + Rand_i^x[U_{\vartheta(x)}^i - L_{\vartheta(x)}^i] \\ \vartheta_{y(i)} = L_{\vartheta(y)}^i + Rand_i^y[U_{\vartheta(x)}^i - L_{\vartheta(x)}^i] \\ \vartheta_{z(i)} = L_{\vartheta(z)}^i + Rand_i^z[U_{\vartheta(x)}^i - L_{\vartheta(x)}^i] \end{cases} \tag{10}$$

$$
\begin{cases}
X(t) = \sum_{i=1}^{n} \vartheta_{x(i)} B_{i,K}(t) \\
Y(t) = \sum_{i=1}^{n} \vartheta_{y(i)} B_{i,K}(t) \\
Z(t) = \sum_{i=1}^{n} \vartheta_{z(i)} B_{i,K}(t)
\end{cases} \quad (11)
$$

$$
\wp^{i}_{x,y,z} = \sum_{i=P^{a}_{x,y,z}}^{|\wp|} \sqrt{(X_{i+1}(t)-X_i(t))^2 + (Y_{i+1}(t)-Y_i(t))^2 + (Z_{i+1}(t)-Z_i(t))^2}
$$

where, the $X(t)$,$Y(t)$,$Z(t)$ display the vehicle's position at each time step $t$ along the generated path $\wp$ in the $\{n\}$ frame. The vehicle's orientation along the generated path also should be calculated as is given by (12):

$$
\psi(t) = \tan^{-1}\left(\frac{|Y_{i+1}(t)-Y_i(t)|}{|X_{i+1}(t)-X_i(t)|}\right)
$$

$$
\theta(t) = \tan^{-1}\left(\frac{-|Z_{i+1}(t)-Z_i(t)|}{\sqrt{(X_{i+1}(t)-X_i(t))^2 + (Y_{i+1}(t)-Y_i(t))^2}}\right) \quad (12)
$$

In (12), the $\psi(t)$ and $\theta(t)$ represent the history the vehicle's yaw and pitch orientations at time $t$. Water currents continually affect the vehicle's motion, so that its position and attitude should be adjusted constantly. The vehicle is assumed to have a constant velocity $\upsilon$. Applying the 3D water current velocity of $V_c$:$(u_c,v_c,w_c)$, given by (13), the vehicle's angular velocity is calculated from the equation (14).

$$
\begin{aligned}
u_C &= |V_C| \cos\theta_C \cos\psi_C \\
v_C &= |V_C| \cos\theta_C \sin\psi_C \\
w_C &= |V_C| \sin\theta_C
\end{aligned} \quad (13)
$$

$$
\begin{aligned}
u &= |\upsilon| \cos\theta \cos\psi + |V_C| \cos\theta_C \cos\psi_C \\
v &= |\upsilon| \cos\theta \sin\psi + |V_C| \cos\theta_C \sin\psi_C \\
w &= |\upsilon| \sin\theta + |V_C| \sin\theta_C
\end{aligned} \quad (14)
$$

The path is evaluated using the following cost function, which is a combination of the time and corresponding penalty functions as described by (15).

$$
\wp(t) = [X(t), Y(t), Z(t), \psi(t), \theta(t), u(t), v(t), w(t)]
$$

$$
T_{\wp} = \sum_{i=1}^{n} \frac{\left|\vartheta^{i+1}_{x,y,z}(t) - \vartheta^{i}_{x,y,z}(t)\right|}{|\upsilon|}
$$

$$
\wp_{Cost} = T_{\wp} + \sum_{i=1}^{n} \gamma_{\nabla} \times f\left(\nabla_{\wp}, \nabla_{\Sigma_{M,\Theta}}\right) \quad (15)
$$

$$
\nabla_{\wp} = [\gamma_u, \gamma_v, \gamma_\theta, \gamma_\psi] \begin{bmatrix} \max(0; u(t) - u_{\max}) \\ \max(0; |v(t)| - v_{\max}) \\ \max(0; \theta(t) - \theta_{\max}) \\ \max(0; |\dot{\psi}(t)| - \psi_{\max}) \end{bmatrix}
$$

$$
\nabla_{\Sigma_{M,\Theta}} = \gamma_{\Sigma_{M,\Theta}} \times \begin{cases} 1 & \wp_{x,y,z}(t) \cap Coast : Map(x,y) = 1 \\ 1 & \wp_{x,y,z}(t) \cap \bigcup_{N\Theta} \Theta(\Theta_p, \Theta_r, \Theta_{Ur}) \\ 0 & Otherwise \end{cases}
$$

In (15), the $\gamma_{\nabla} \times f(\nabla_{\wp}, \nabla_{\Sigma M,\Theta})$ is a weighted violation function including AUV's kinodynamic constraints and collision boundaries. The generated path shouldn't have any intersection with the forbidden zone ($\Sigma_{M,\Theta}$) associated with coast sections of the map and obstacles $\Theta$. The coastal area are determined applying a k-means clustering method on offline map, which more detail is provided in section 4. The $\gamma_u$, $\gamma_v$, $\gamma_\theta$, $\gamma_\psi$, and $\gamma_{\Sigma M,\Theta}$ are impact of each violation in calculation of the $\wp_{Cost}$.

### *On-line Path Re-planning based on Previous Solution*

This research applies on-line path planning as a continuous process for correcting the previous trajectory, so that the vehicle can be able to take a prompt reaction to cope with immediate changes of the dynamic environment. To this principle, environmental situation should be simultaneously computed and consequently the current trajectory should be refined continually during deployment toward the destination. To accomplish the re-planning, information of the vehicle's current position and velocity along with dynamic changes of the environment are fed to the to the LOP-P module. In the proposed mechanism when the path

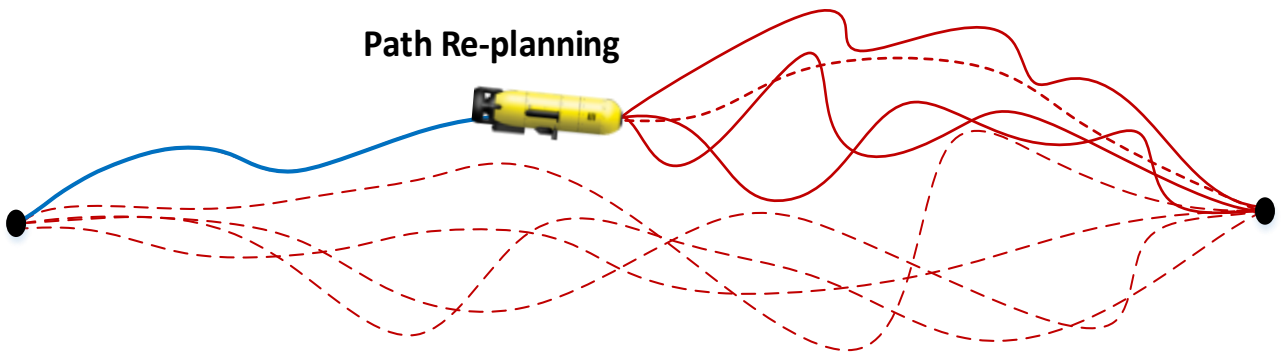

**Online Path Re-Planning Mechanism**

- Get the position of start and target points from the TOM-P module
- Generate the initial optimum path using Eq.(11)-(14)
- Evaluate the generated path by defined $\wp_{Cost}$ in Eq.(15)
- Check Path-Replanning Flag continuously

**BEGIN**

  ***If*** *Flag ==1*

    –Apply the current states of the vehicle as the initial condition

    –Set the current location as the start point for LOP-P process

    –Use the solution at hand (from previous path) as an initial solution or initial guess for the proposed evolutionary method

    –Generate a new path

    –Check the problem optimality conditions Eq.(15)

    ***If*** Eq.(15) is satisfied

      Follow the new path

      Terminate the planning process

      ***Otherwise***

      Follow the pervious path

      Terminate the planning process

      Back to **BEGIN**

    ***End***

  ***End***

**END**

**Fig.3.** On-line path re-planning mechanism

re-planning flag is triggered the LOP-P generate a new alternative path from the vehicle's current position to the target point by rectifying the previous trajectory. In other words, instead of regenerating the path from beginning in the problem search space, it would be more appropriate if the old optimum path get refined from the existing position to the destination, which is discussed more in analysis of the simulation results. The *Fig*.3 clarifies this issue. In this fashion, the vehicle would be able to accurately cope with dynamic variations and uncertainty of the underlying environment. Indeed, this mechanism is resembling closed-loop guidance configuration and also reduces the computational cost [24]. This process is summarized in Fig.3.

## 2.3 Modelling of the Synchron Module

After the TOM-P generated a mission plan in a sequence of tasks/waypoints, the LOP-P module gets this sequence along with environmental information and starts generating safe and efficient trajectory between the waypoints while the AUV concurrently executes the corresponding tasks. In this way the AUV efficiently carries out its mission and safely guided through the waypoints. The provided trajectory is shift to the guidance controller to produce the guidance commands. During deployment between two waypoints, the LOP-P can incorporate any dynamic changes of the environment. The LOP-P repeatedly calculates the trajectory between vehicles current position and its specified target location (any time that new update of information received from the sensors). After each waypoint is visited the total path time of $T_\wp^{ij}$ gets compared to expected time $T_\varepsilon^{ij}$ for traversing the corresponding distance, in which the $T_\varepsilon^{ij} \approx t_{ij}$ is extracted from the mission time $T_M$ given by (2). In a case that the $T_\wp^{ij}$ gets smaller than $T_\varepsilon^{ij}$, the current order of tasks would be valid and the vehicle can continue its mission. But in the opposite case, when the $T_\wp^{ij}$ exceeds the $T_\varepsilon^{ij}$, it means a certain amount of the available time $T_{Total}$ is spent for handling any probable challenge during the path planning process. In such a case, the previously defined mission scenario cannot be optimum anymore; therefore, mission re-planning according to mission updates would be necessary.

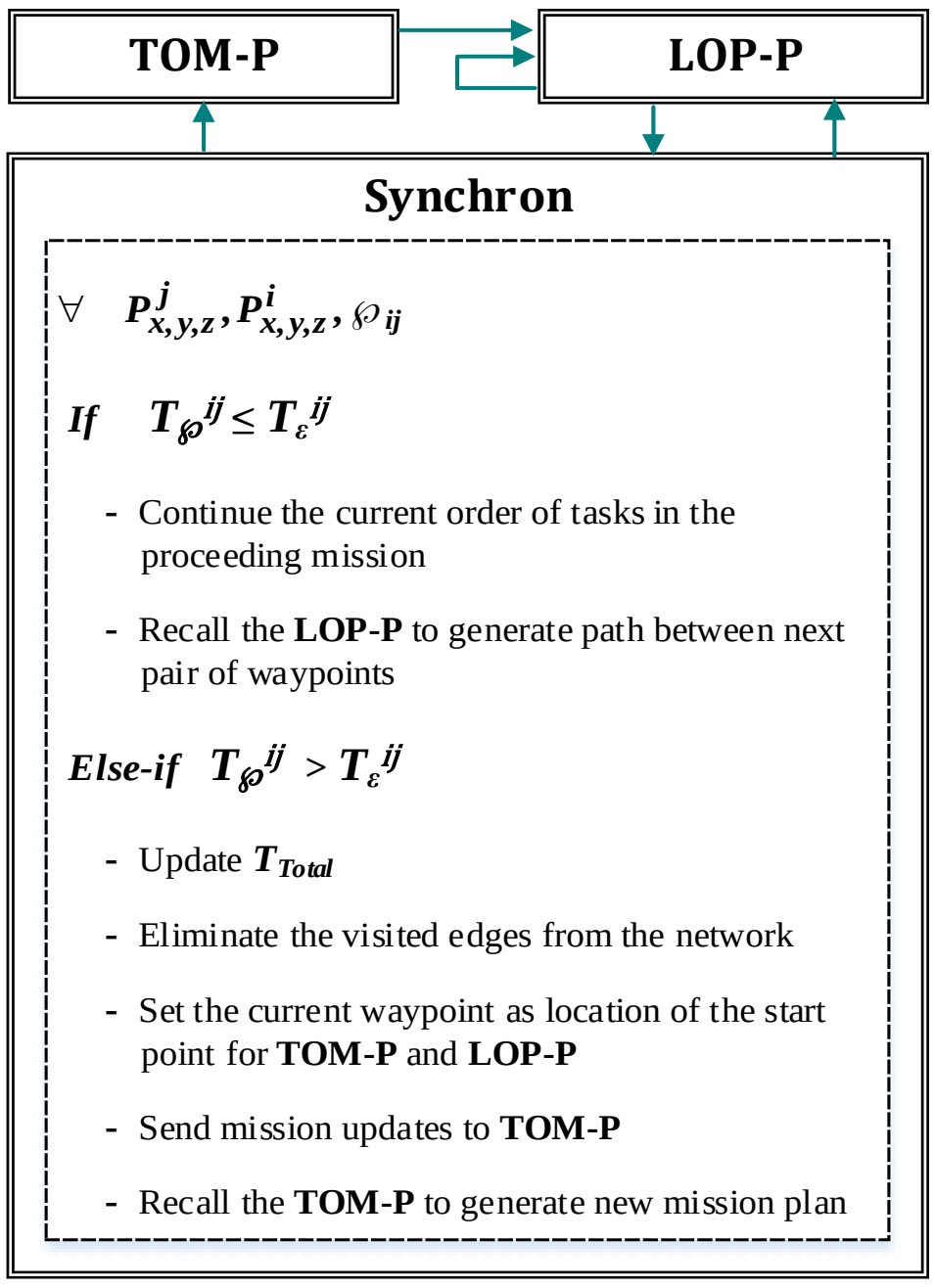

**Fig.4.** The synchronization process

However, a question is raised in this situation about computational burden of the mission re-routing process. Passing a specific edge (distance) for more than ones is the time dissipation for an AUV that means repeating a task for several times. To prevent this issue, the following steps should be carried out any time that re-routing is required: the $T_{Total}$ gets updated; the passed edges get eliminated from the operation network (so the search space shrinks); and the location of the present waypoint is considered as the new start position for both LOP-P and TOM-P. Afterward, the TOM-P tends to plan a reorder the tasks and prepare a new mission scenario based on new information and updated network topology. A computation cost encountered any time that mission re-planning is required. This process continues until vehicle reaches the required waypoint. The trade-off between total available time and mission objectives is critical issue that should adaptively carried out by TOM-P. Hence, the main architecture should be fast enough to track environmental changes, cope with dynamic changes, and carry out prompt path or mission re-planning based on raised situation. The proposed synchronous architecture gets evaluated according to criterion given in the next section.

## 2.4 Architecture Evaluation Criterion

Performance of the generated local path by LOP-P is evaluated by a cost function $\wp_{Cost}$ based on travel time ($T_\wp$), which is proportional to energy consumption and the distance travelled. The mission cost $M_{Cost}$ has direct relation to the passing distance between each pair of selected waypoints. Hence, the path cost $\wp_{Cost}$ for any optimum local path gets used in the context of the TOM-P. The model is seeking for an optimal solution in the sense of the best combination of path and mission cost as follows:

$$\begin{aligned}
&\forall \wp^{j}_{x,y,z} = [X(t),Y(t),Z(t),\psi(t),\theta(t),u(t),v(t),w(t)]\\
&\wp_{Cost} \approx T_\wp = \sum_{i=P^{a}_{x,y,z}}^{|\wp_{x,y,z}|} \frac{\sqrt{(X_{i+1}-X_i)^2+(Y_{i+1}-Y_i)^2+(Z_{i+1}-Z_i)^2}}{|\upsilon|}\\
&s.t.\\
&[X(t),Y(t),Z(t)] \cap \Sigma_{M,\Theta} = 0\\
&u(t) \le u_{\max} \quad \& \quad v_{\min} \le v(t) \le v_{\max}\\
&\theta(t) \le \theta_{\max} \quad \& \quad \dot{\psi}_{\min} \le \dot{\psi}(t) \le \dot{\psi}_{\max}
\end{aligned} \tag{16}$$

$$\begin{aligned}
&M_{Cost} \propto |T_M - T_{Total}| = \left| \sum_{\substack{i=0\\ j\neq i}}^{n} lq_{ij} \times \left(\wp^{ij}_{Cost} + \wp_{\mathrm{CPU}}\right) - T_{Total} \right| + \sum_{\substack{i=0\\ j\neq i}}^{n} \frac{lq_{ij}}{w_{ij}}; \quad l \in \{0,1\}\\
&s.t.\\
&\max(T_M) < T_{Total}
\end{aligned} \tag{17}$$

After visiting each waypoint, the re-planning criterion (given by *Fig*.4) is investigated. A computation cost encountered any time that re-routing is required. Thus, the total cost $Cost_{Total}$ for the model is defined as:

$$Cost_{Total} = M_{Cost} \times f(\wp_{Cost}) + \sum \wp_{\mathbf{CPU}} + \sum_{1}^{rep} T_{compute} \times (M_{\mathbf{CPU}}) \qquad (18)$$

where *rep* is the number of repeating the mission re-planning procedure.

## 3 Methodology Adopted by Higher/Lower Level Modules of the Architecture

There is a significant distinction between theoretical understanding of meta-heuristics and peculiarity of different applications in contrast. Specifically this gap is more highlighted when scale (size), complexity, and nature of the problem is taken to account. Application of different methods may result very diverse on a same problem due to specific nature each problem. To handle the complexity of NP-hard graph routing and task allocation problem, the TOM-P module utilizes the Ant Colony Optimization (ACO) algorithm to find an optimum order of tasks for the underwater mission. In the LOP-P module, Firefly Algorithm (FFA) is conducted to carry out path planning between each pair of the waypoints, which is efficient and fast enough in generating collision-free optimum trajectory in smaller scale. Similar to other metaheuristic algorithms, both ACO and FFA also have two inner loops through the population size $i_{max}$ and iteration $t_{max}$, so at the extreme case the algorithms complexity is $O(i_{max}^2 \times t_{max})$; thus, the computation cost is considerably low as its complexity is linear to time.

### 3.1 Application of ACO Algorithm on Task-Organize Mission Planning Approach

The actual biological ants adopts swarm intelligence to discover the shortest possible route to food. The ACO algorithm has been introduced by Dorigo & Di Caro [37] inspiring the ants' foraging process. This algorithm is proven to have a strong capability on solving vehicle routing problems and is well scaled with task assignment approaches due to its discrete nature [38-40]. In this algorithm, the ants find their path through a probabilistic decision according to level of pheromone concentration, so that the pheromone trail attract ants and guide them to grain source. Respectively, more pheromone is released on the overcrowded paths, in which produces a self-reinforcing process in finding the quickest route to target. The pheromone gets vaporized over time to helps ants to eliminate bad solutions that keep solutions from falling in to local optima. However, in evaporation process the pheromone concentration on optimum path still remains higher due to lower rate of evaporation comparing to pheromone deposit rate. The pheromone concentration on edges (τ) preserve the ants' collected information in searching process. For implementing ACO two main steps are required, first determination of the probability of targeting a node to travel among other nodes, which is calculated by (19), and the next is to update the pheromone trail after ant population completed their tour that is given by (20).

$$p_{ij}^{k} = \begin{cases} \dfrac{\left(\tau_{ij}^{(k)}\right)^{\alpha}\left(\Phi_{ij}^{(k)}\right)^{\beta}}{\sum_{l \in N_l^k}\left(\tau_{il}^{(k)}\right)^{\alpha}\left(\Phi_{il}^{(k)}\right)^{\beta}} & j \in N_i^k \\ 0 & j \notin N_i^k \end{cases} \qquad (19)$$

$$\tau_{ij} = (1-r)r + \sum_{k=1}^{N} \Delta\tau_{ij}^{(k)}$$
$$\tau_{ij}(t+1) = \tau_{ij}(t) + \Delta\tau_{ij} \qquad (20)$$
$$\Delta\tau_{ij} = \begin{cases} -\lambda\tau_{ij}(t) + \dfrac{Q}{\Omega} & \text{For nodes of the best ant} \\ -\lambda\tau_{ij}(t) & \text{Otherwise} \end{cases}$$

```
ACO Based Mission Planning
Initialization phase:
  – Set the maximum number of iterations t_max
  – Set the maximum number of ants population i_max
  – Define number of layers in ant motion network, where each layer contains p nodes
    i.e. (χ_i1, χ_i2, …, χ_ip).
  – Set initial pheromone for each edge by τ_ij^(l)=1
  – Initialize ants in a colony with randomly generated feasible task sequences

For t =1: t_max
    For k =1: i_max
        Compute the probability P_ij of selecting node j from the current node i by (19).
        ▪ Assume α=1.
        ▪ Generate k random numbers r_k∈[0,1], one for each ant.
        ▪ Calculate range of the cumulative probability associated each routes applying
          roulette-wheel selection.
        ▪ Determine the route captured by ant k as a cumulative probability that keeps
          r_k random number in it.
        ▪ Evaluate the cost value for all solutions M_Cost^k=M_Cost(x_t^(k)); k=1,2,…,n and
          determine best and worst solutions x_t^best, x_t^worst
        ▪ Check the solution feasibility criterion
            [1] Sequence should be commenced with index of the start waypoint.
            [2] Sequence should be ended with index of the target waypoint.
            [3] A task on a specific edge should not be repeated more than once in a sequence.
            [4] Sequence should not include nonexistent edges in the network.
        ▪ Save the best solution x_t^best
        Update the pheromone on arc when all the ants complete their tour by (20).

        For j =1:k
            τ_ij^k = τ_ij^old + Σ_{k=1}^{i_max} Δτ_ij^k
            τ_ij^old = (t − ρ)τ_ij^(t−1)
                If M_Cost(x_t^best)^new < M_Cost(x_t^best)^old
                    x^Globalbest=(x_t^best)^new
                else
                    x^Globalbest=(x_t^best)^old
                end
        end
    end
    α_t =0.99×α_(t-1)
    β_t =0.99×β_(t-1)
end
Output result
```

**Fig.5.** Pseudo-code of ACO mechanism on task organizing

here, $k$ is the ant index, $i$ and $j$ denote the index of current and the target nodes, $\tau^{k}_{ij}$ is the pheromone concentration on edge between nodes $i$ and $j$, α and β are the pheromone and heuristic factors, respectively. $\Phi_i$ denotes the heuristic information in $i^{th}$ node. $N_i^{(k)}$ indicates the set of neighbor of ant $k$ when located at node $i$. The $r \in (0,1]$ is evaporation rate. The $\Delta\tau_{ij}^{(k)}$ denotes the pheromone level collected by the best ant $k$ based on priority of ants released pheromone (Ω) on the selected nodes. $\tau^{k}_{ij}(t)$ and $\tau^{k}_{ij}(t+1)$ are the previous and current level of the pheromone between node $i$ and $j$. $Q$ is the amount of released pheromone on nodes taken by the best ant(s), and λ is the coefficient of pheromone evaporation. The procedure of the ACO based mission planning is described by *Fig*.5. Initializing the ant population is the most critical step in implementing ACO framework. In this research, the ant population initialized with feasible waypoint sequence using a random priority vector and adjacency information of the operation graph (for more detail refer to [30]). The feasibility of the generated solutions is checked during the process of cost evaluation.

## 3.2 Application of FFA on the Path Planning Approach

The FFA is another swarm-intelligence-based meta-heuristic algorithm introduced by Xin-She Yang [41,42]. This algorithm inspired from the flashing patterns of fireflies, in which the fireflies attracted to each other based on brightness and regardless of their sex. As the distance of fireflies increases their brightness gets dimmed. The less bright firefly approaches to the brighter one. Attraction of each firefly is proportional to its brightness intensity received by adjacent fireflies and their distance *L*. The distance of *L* between two fireflies of $\chi_i$ and $\chi_j$ can be defined as follows:

$$L_{ij} = \left\|\chi_j - \chi_i\right\| = \sqrt{\sum_{q=1}^{d} \left(x_{i,q} - x_{j,q}\right)^2} \qquad (21)$$

where $x_{i,q}$ represents the $q^{\text{th}}$ component of the firefly $\chi_i$ coordinate in d dimensions. The attraction factor $\beta$ and movement of a firefly $\chi_i$ toward the brighter firefly $\chi_j$ is calculated by

$$ß = ß_0 e^{-\varepsilon L^2}$$
$$\chi_i^{t+1} = \chi_i^t + ß_0 e^{-\varepsilon L_{ij}^2}(\chi_j^t - \chi_i^t) + \alpha_t \varsigma_i^t \qquad (22)$$
$$\alpha_t = \alpha_0 \kappa^t, \quad \kappa \in (0,1)$$

where $\beta_0$ is the attraction value at $L$=0, $\alpha_t$ is the randomization parameter that control the randomness of the movement and can be tuned iteratively. The $\alpha_0$ is the initial randomness scaling value and $\kappa$ is the damping factor. The $\varsigma_i^t$ is a random vector generated by Gaussian distribution at time *t*. There should be a proper balance between engaged parameters, because if the $\beta_0$ approaches to zero, the movement turns to a simple random walk, while ε= 0 turns it to a kind of particle swarm optimization [41]. The mechanism of the algorithm propels the firefly *i* at $\chi_i$ toward the attractive firefly *j* at $\chi_i$ using:

```
Initialization phase:
 – Initialize fireflies χq (q=1,2,…,n) with the control points (ϑ) of the candidate path ℘i
 – Define light absorption coefficients ε
 – Initialize the attraction coefficient β0
 – Set the damping factor of κ
 – Initialize the randomness scaling factor of α0
 – Set the parameter of randomization αt
 – Set the maximum iteration tmax
 – Set the number of population imax
For t =1: tmax
  For i =1:imax
    Reconstruct a path according to χi vector
    Evaluate the path
    Update light intensity of χi
    For j =1:i
      Reconstruct a path according to χj
      Evaluate the path
      Update light intensity of χj
        If (βj > βi),
          Move firefly i towards j
        end
    end
  end
  Rank the fireflies and find the current best
end
Output result
```

**Fig.6.** FFA mechanism on path planning approach

$$\chi_i = \chi_i + ß_0 e^{-\varepsilon L_{ij}^2}(\chi_j - \chi_i) + \alpha[rand - \frac{1}{2}] \qquad (23)$$

where, *rand* is for generating uniformly distributed random number in [0,1]. For more detail about FFA refer to [40]. The FFA is advantaged uses an automatic subdivision approach that makes it more efficient comparing to other optimization algorithms. The attraction decreases with distance and in this context entire population automatically get subdivided, where each subgroup swarm around its local optimum and the best global solution is selected among the set of the local optimums. This fact increases convergence rate of the algorithm. Such an automatic subdivision nature motivates fireflies to find all optima iteratively and simultaneously, which makes the FFA specifically suitable and flexible in dealing with continuous problems, highly nonlinear problems, and multi-objective problems [43,44]. The control parameters in FFA can be tuned iteratively that increases convergence rate of the algorithm too. To generate path by B-Spline curves, the FFA tends to accurately locate the control points of a candidate curve in the solution space. A firefly in this context corresponds to a candidate solution (path) involving a distinct number of control points. The pseudo-code of FFA process on path planning approach is provided by *Fig*.6.

# 4 Discussion and Analysis of Simulation Results

To validate the architecture's performance, first higher and lower level modules are evaluated separately, then their synchronous collaboration is investigated and evaluated.

## 4.1 Simulation Results of LOP-P Performance in Small Scale Dynamic Environment

The purpose of the LOP-P module in this study is to find the efficient trajectory adjustments making use of desirable current field, escaping undesirable current flows, and avoid intersecting uncertain borders of static/moving obstacles. Terrain situation and updates get observed and continuously tracked using on-board sensors, such as a Horizontal Acoustic Doppler Velocity Logger. A real map data of Adelaide Vincent gulf [33] is utilized to model more realistic marine environment. A k-means clustering algorithm is employed where the *k* (number of clusters) is set by 3 to cluster there sections of coast, water zone, and uncertain areas of the map and transform it to matrix form where the blue sector of the image is sensed as water and defined as permissive area for vehicle's operation. Other sections depending on their color range categorized as impassable and uncertain zones. The utilized clustering algorithm is highly efficient in recognition of mentioned sectors in any alternative map. The AUV water-referenced velocity is set on 2.5 m/s. Number of control points for generating B-spline curves is set to alter between 5 to 8 points. The common optimization configuration of FFA is set with 100 iterations and population of 80 fireflies. The attraction coefficient base value is set on $\beta_0$=2, light absorption coefficient ε is assigned with 1. The damping factor of $\kappa$ is assigned with 0.95 to 0.97. The scaling variations is defined based on initial randomness scaling factor of $\alpha_0$. The parameter of randomization is set on 0.4. The simulation results for three different cases are presented by *Fig*.7 to *Fig*.11, in which the terrain gets more complex at each case. Efficiency of the proposed method is investigated at the most sever situation that closely matches the real underwater dynamics. A simple path planner and On-line path re-planner are compared at each case to declare the necessity for existence of an adaptive re-planning procedure in such a dynamic environment.

### A. *Path Adaptive Behavior to Dynamic Current Variations (without obstacle)*

Water current may have positive or disturbing effect on vehicles deployment. An undesirable current also can push floating obstacles across the AUVs trajectory. On the other hand, the desirable current can motivate AUVs motion toward its target, which leads saving the energy. Adapting to dynamic current is even more challenging where the current flow may immediately change its magnitude or correlation. The real-time current map information is continuously captured by on-board sonar sensors with a specific uncertainty modelled by Gaussian distribution (formulated in section 2.2.1). Vehicle's kinematic constraints also are taken into account in path planning process and investigated in the following.

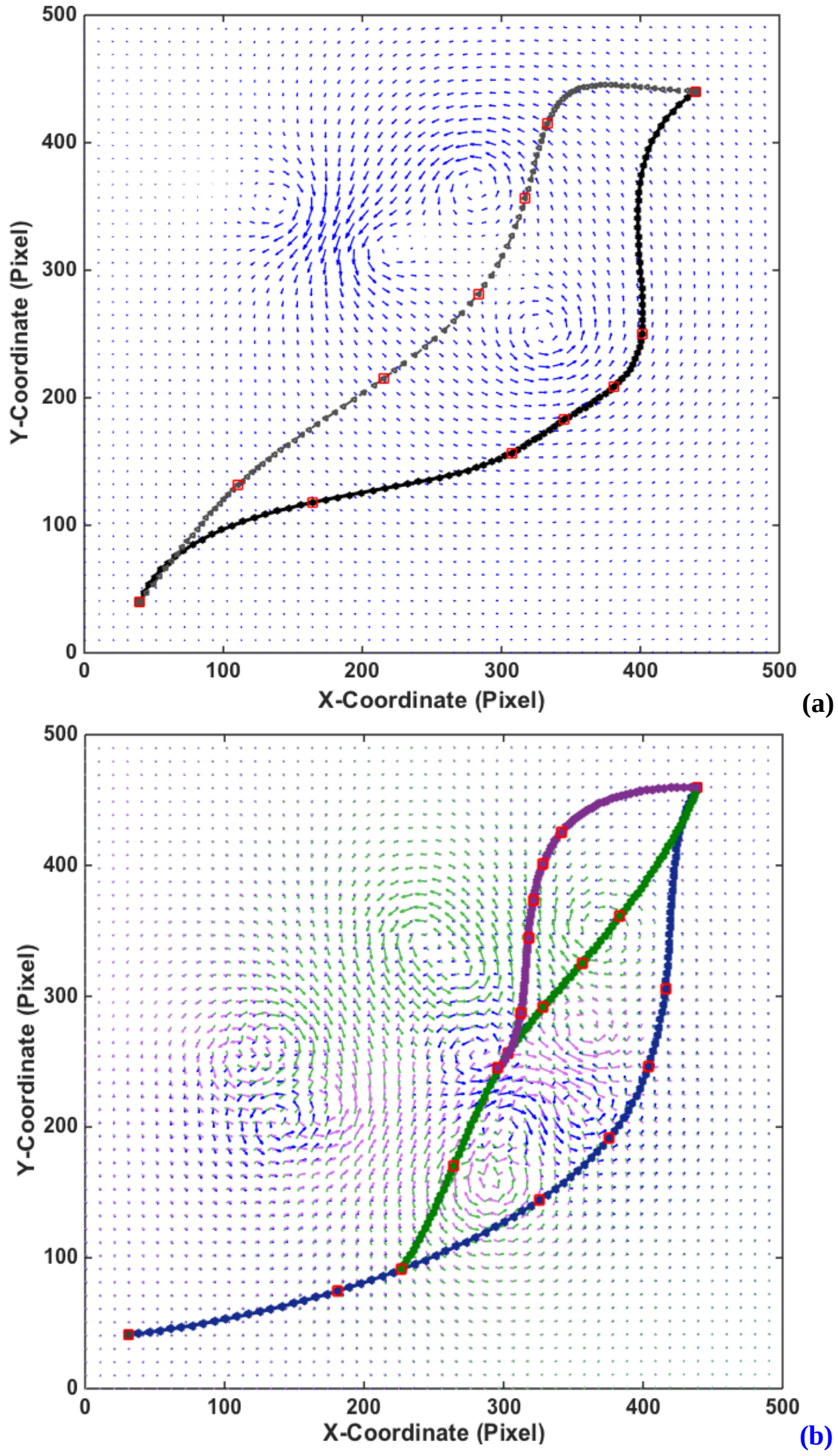

**Fig.8.(a)** Path curve and its alternative generated by simple path planner encountering information of the static current map; **(b)** The path adaption to current updates in three levels of On-line replanning. Each pixel corresponds to 7×7 $m^2$.

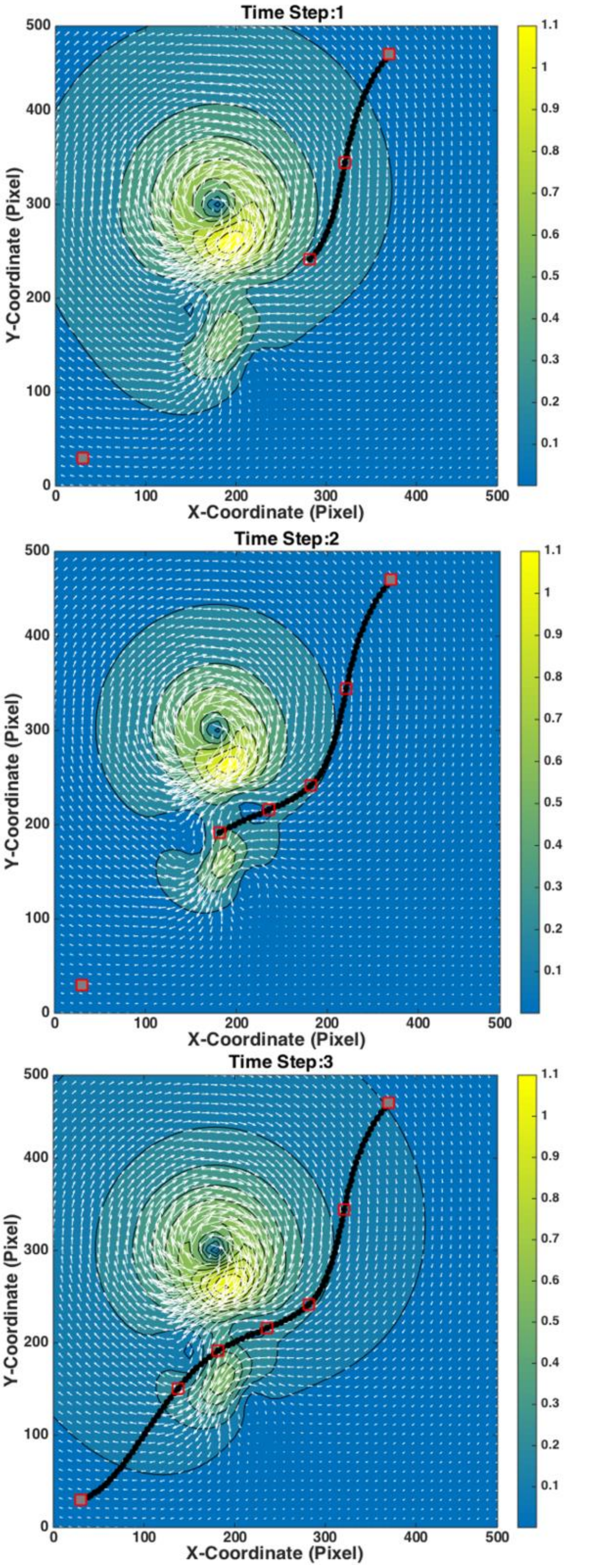

**Fig.7.** Path adaption to current updates in three time steps (the color map corresponds to current magnitude. The current intesity increases as the colour gets lighter from blue to yellow). Each pixel corresponds to 7×7 $m^2$.

The behaviour of the generated path to current updates is proposed in *Fig*.7 and *Fig*.8(a,b), in which the current map gets updated in three time steps where time interval for current update is determined by division of the number of iterations to current update rate ($U_R^C$=3) that is set and initialized by the operator. Size of the map used in LOP-P is set on 500×500 pixels and the current field computed from a random distribution of 3 to 6 Lamb vortices in 500×500 grid; hence, each pixel in *Fig*.7 and *Fig*.8 contains a current arrow and corresponds to 7×7 $m^2$. *Fig*.7 presents the path curve deformation in three step current update in which the colour map presents the current magnitude and the current intensity increases as the arrows approach to the yellow areas. It is noted from *Fig*.7 that the generated path accurately avoids the higher intensity adverse current arrows; more specifically, it is clear in second and third time steps that the path curve is mounted on accordant current arrows to take energy efficient distance to the target point. For better presentation *Fig*.8 (a,b) is provided to show accuracy of the generated path by LOP-P module in which current updates and re-planned paths are presented by the same colour. In *Fig*.8(b) the blue arrows correspond to first current map in which the generated primary path is drawn with the same colour accordingly; afterward, the second current map arises that presented by green arrows, where the path planner tends to correct the primary path (blue) from the existing location to the target point in accordance with second current map presented with green line. In the third update, the On-line path planner again takes the accurate trajectory to the destination in accordance with third current map, where both corrected trajectory and last updated current map are presented with dark purple colour. Considering the path deformation it is noteworthy to hint the efficiency of the proposed method in adapting prompt current changes whether in avoiding undesirable turbulent or driving the desirable flows.

### B. Path Adaptive Behavior in Complex Environment Including Dynamic Current and Uncertain Static-Moving Obstacles

To test the performance of the adopted method by LOP-P, the operating field is modelled to be more complex encountering static obstacles (presented by black circles) and static current (given in *Fig*.9(a)) and uncertain moving obstacles in presence of dynamic current (given in *Fig*.9(b)). In *Fig*.9 a set of pareto-optimum trajectories are generated in accordance with environmental situation in which the best generated path is proposed by thicker line and the sub optimum trajectories are presented by a thinner lines. In *Fig*.9(b) current map is updated for three times and the path re-planned adaptively at each current update. Further to current update, the uncertain obstacles are also taken into account, in which some of them are static and uncertain in position where their uncertainty grows by time in a circular format; and some others are floating with current force with uncertainty on their position and motion (as the current flow also contains uncertainty) that presented by moving circles with varying radius. Similar to previous case, each pixel in *Fig*.9 contains a current arrow.

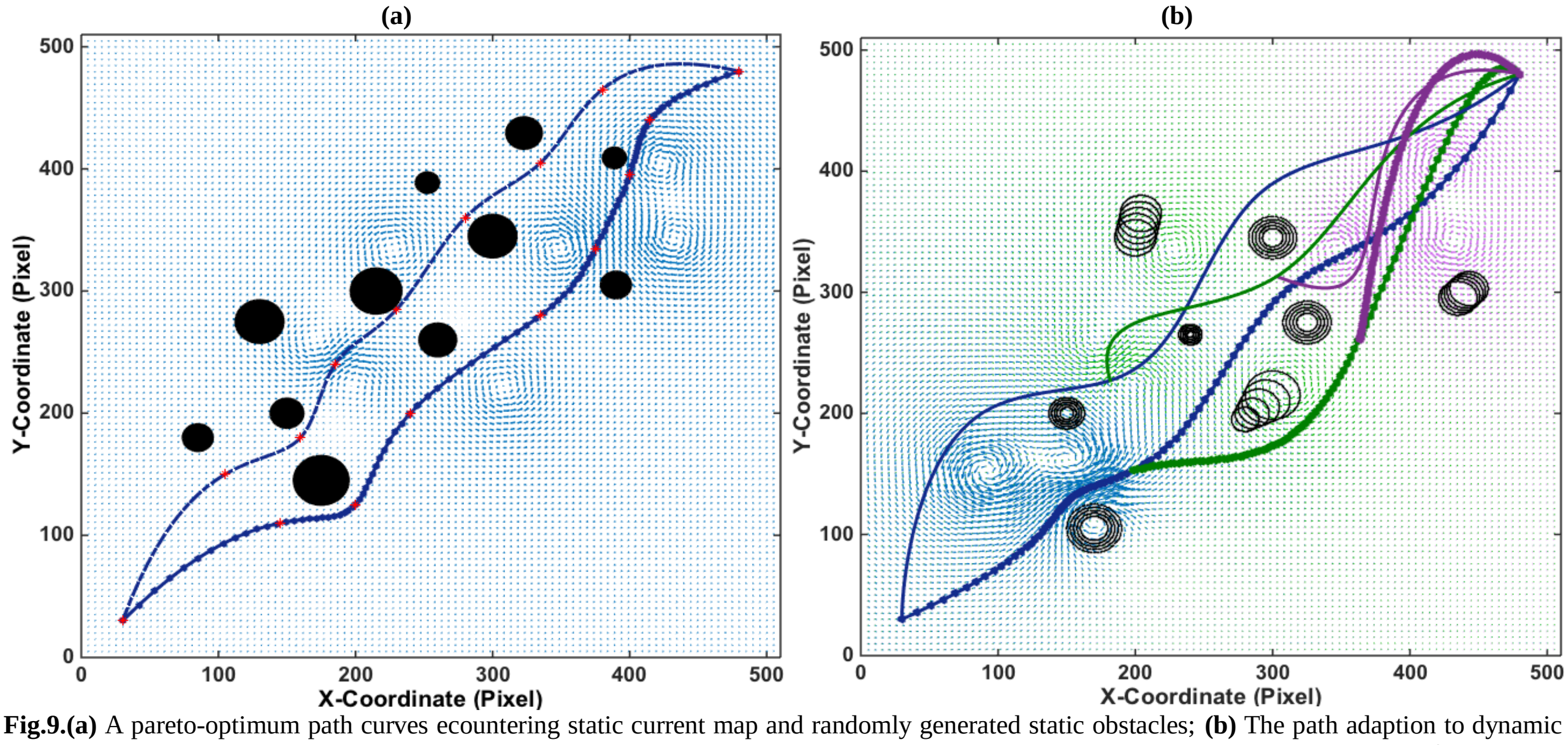

**Fig.9.(a)** A pareto-optimum path curves ecountering static current map and randomly generated static obstacles; **(b)** The path adaption to dynamic current and uncertain static/moving objects along with On-line replanning, where the current updates and the corresponding path deformation is presented in the same colours. Each pixel corresponds to 7×7 $m^2$.

It is derived from simulation results in *Fig*.9(a) the proposed simple path planner accurately handles the problem's objectives in a static situation based on offline static current map data; however, for handling more complex situations adaptive re-planning would be necessary to correct the previous optimum trajectory according to terrain updates. As presented in *Fig*.9(b), the On-line re-planning accurately adapts the changes of the environment as the primary blue trajectories that generated based on first current map(blue arrows) are switched to green lines from the existing point to the target point in accordance with second current map(green arrows) and obstacles movement/deformation. The purple lines are the final correction to the trajectories based on latest update of the current map (purple arrows) and obstacle's situation. Obviously, the employed strategy satisfies the path planning objectives at this stage; however, for gaining more confidence on efficiency of the applied method, the terrain gets even more complex to closely mimic the real-world underwater terrain that investigated in the next section.

### C. On-line Path Planning in 3-D Complex Environment Encountering Real Map Data, Dynamic Current and Obstacles

Similar to case (B), both static and dynamic ocean currents are applied and different type of obstacles are considered for final evaluation of the adopted method by LOP-P module. To be more confident on accuracy of the proposed method, a part of a real map also adopted for more realistic implementation and k-means clustering is applied to clarify water covered zone as permissive space for vehicle's deployment and coastal areas as the forbidden zones and transform it in to a matrix format, in which the water covered sections filled by value of 1 in the generated matrix and the other sections filled by zero for coastal area or a small decimal value in range of [0.01, 0.3] corresponding to recognized uncertain sections of the map.

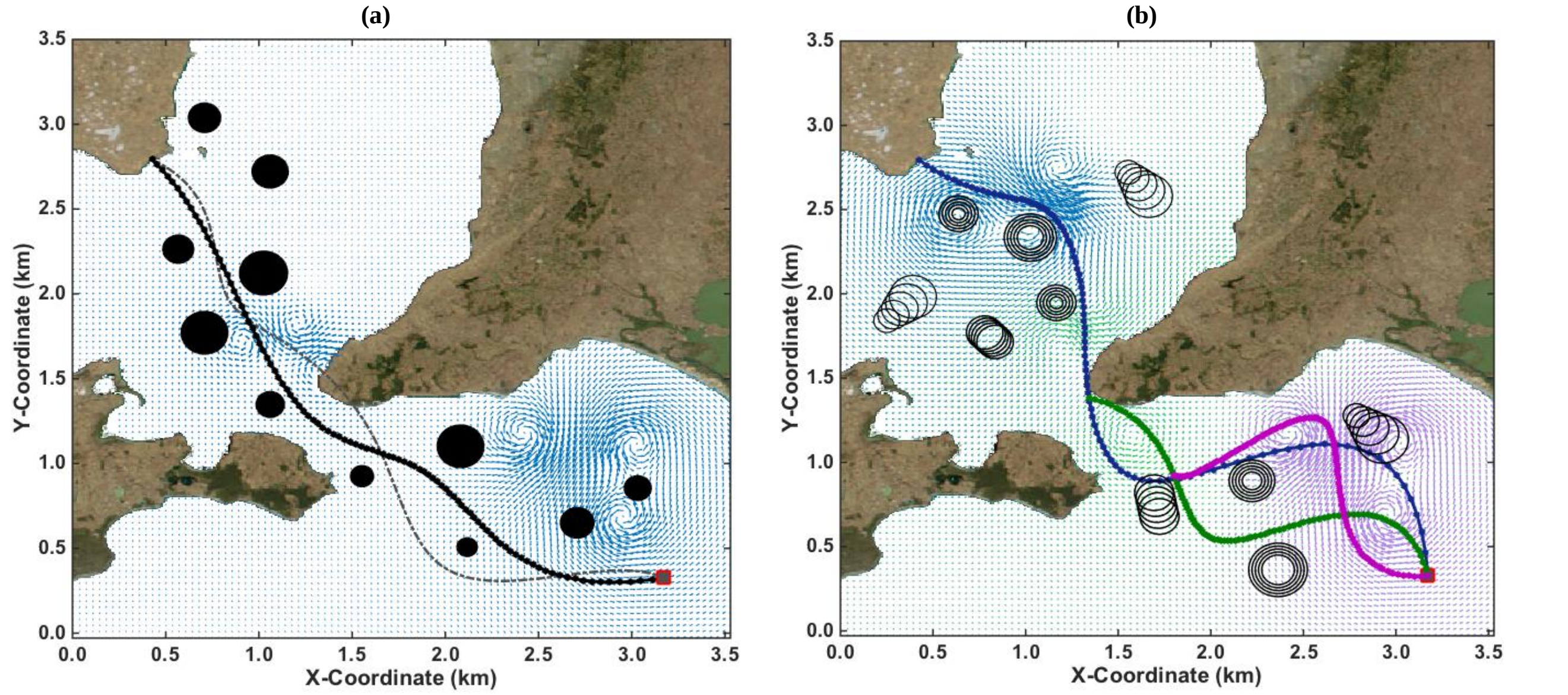

**Fig.10. (a)** Optimum and alternative sub-optimum path curves ecountering real map data, static current map, and randomly generated static obstacles; **(b)** On-line path replanning to adapt dynamic current and behaviour of the uncertain obstacles, where current updates and corresponding path deformation are presented in three colours (blue, green, purple).

It is clear from *Fig*.10(a,b) the path planner accurately recognizes coastal areas and avoid colliding forbidden edges. Similar to results performed in previous sections the method is resistant to increasing the complexity of the terrain as it is accurately guide the vehicle toward the target point even in presence of time-varying current and uncertain static/moving obstacles. The path accurately reformed from existing point according to current update. For better presentation and easy understanding, each deformation to the path curve is presented by the same color of the updated current map. Considering *Fig*.10(b) the deformed path at each replanning process accurately copes the new current map and takes the most accordant trajectory to the desirable current flow that leads considerable save to the battery usage. On-line path re-planning uses the previous planning history to achieve more optimized path and save the computation time.

***Path Efficiency in Satisfying Vehicle's Kinematic Constraints Incorporating Current Effects on AUV's Motion***

The vehicular constraints of the AUV and boundary conditions on state of the vehicle also are taken into account for realistic modelling of the AUV's operation. In fact, the optimum trajectory is obtained thru the adjustments of the path curvature considering the current effects on vehicles motion and collision boundaries. The performance of the applied method in satisfying vehicular constraints is presented by *Fig*.11 using compact boxplots over 100 iterations. The displayed four parameters present vehicle's orientation and velocity components at each single spot on the generated path curve. The red dashed line in each plot represents the violation boundary, while the blue boxes corresponds to average iterative variations of the surge, sway, yaw and theta parameters of corresponding path population. The results presented by *Fig*.11 reveals that this model efficiently keeps the variations of the vehicular parameters valid to the specified boundaries as all generated solutions for all four aforementioned parameters are bundled between defined borders (red horizontal lines).

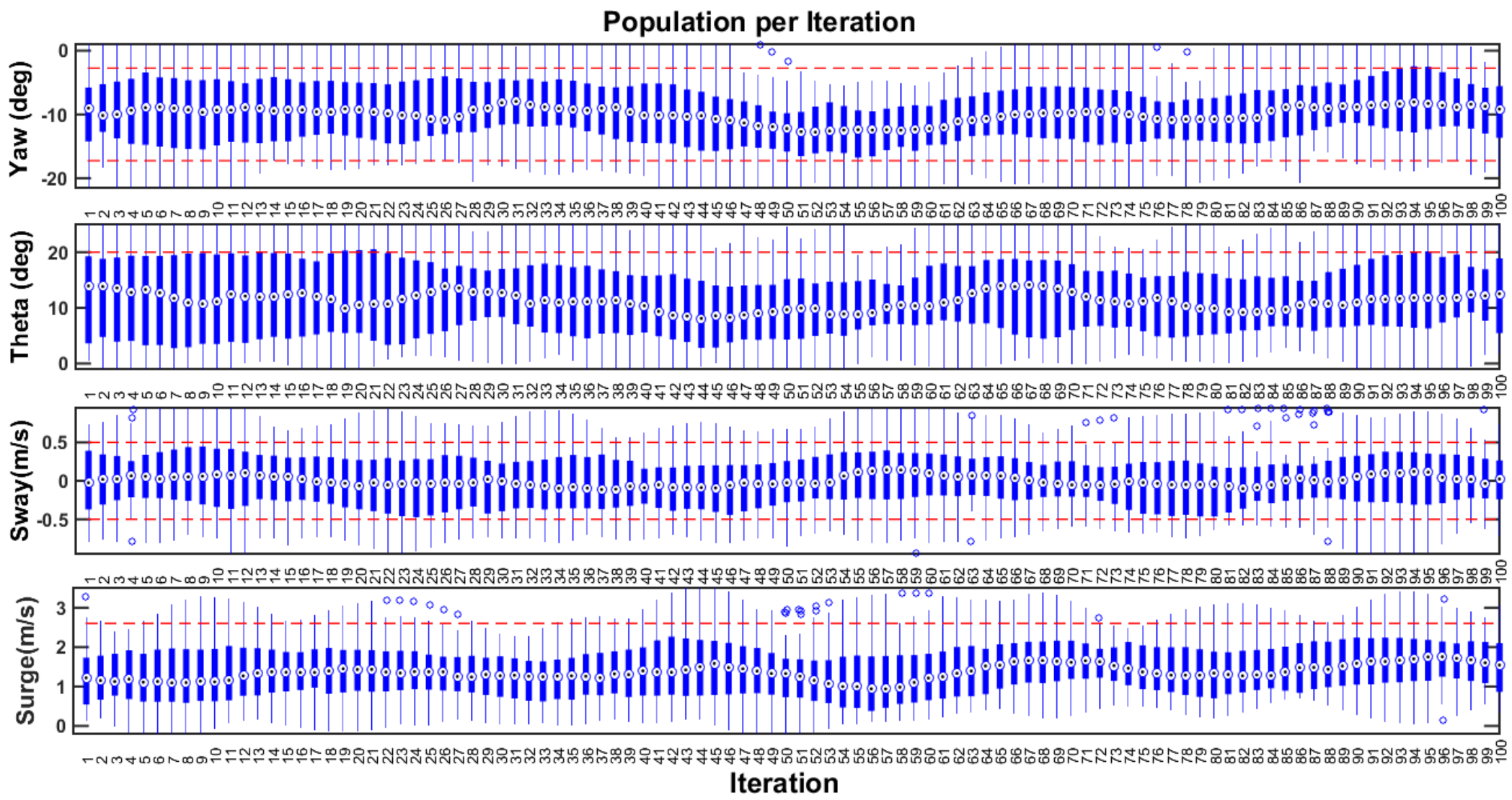

**Fig.11.** Average iterative variations of Yaw, Theta, Sway, and Surge parameters of the path population.

The shortcomings with the LOP-P raises when vehicle is required to operate in a large scale terrain (i.e 10 $km^2$ or even more), as it requires to consider and estimate dynamicity of the terrain adaptively. Hence, when the scale is larger the data load for adaptive re-planning progressively increase, so that the computation time extremely increases, which is a considerable defect for real-time system. Moreover, path planner is an estimate based approach and it is far from reality to give proper estimation on current or terrain behavior in hundreds of meters away from the vehicle. On the other hand this approach only deals with vehicles guidance from one point to another and does not deal with mission scenario or task assignment considerations. To cover the mission scenario and tasks priority assignment and also to handle the shortcomings of the local path planning, the TOM-P operates in a higher level to give a general overview and cut off the operating area to beneficial zones for vehicles deployment in the feature of the global route (waypoint sequence).

## 4.2 Simulation Results of TOM-P Performance on Task Organize and Mission Plan in a Large Scale Graph-like Terrain

The TOM-P module in this research uses prepared information about terrain, waypoints location, tasks that assigned to network's adjacent connections and their characterization, and time/battery restrictions to compute the most appropriate order of prioritized tasks to guide the AUV toward its destination in a waypoint covered large scale terrain. There should be a compromise among the mission available time, maximizing number of highest priority tasks, and guaranteeing reaching to the predefined destination before the vehicle runs out of battery/time. The ACO algorithm is employed and tested by TOM-P module to evaluate its stability and performance in determining efficient feasible solutions and beneficent ordering of the tasks. Several performance metrics are utilized to measure the performance of the method in a quantitative manner and the simulation results presented by *Fig*.12 to *Fig*.13. Also different criteria are embedded to keep the generated solutions concentrated to the feasible space. The ACO is configured with 100 iterations, ant population of 70. The pheromone value initialized with $\tau_0$=1.5, the pheromone exponential and heuristic exponential weights are initialized with 1.5 and 0.9, respectively. The pheromone and heuristic factors are updated iteratively according to $\alpha_t=0.99\times\alpha_{t-1}$ and $\beta_t=0.99\times\beta_{t-1}$. The evaporation rate is assigned with 1.8. The performance and stability of the applied method is investigated on different network topologies, in which network complexity and size increases incrementally from 20 to 150 nodes (waypoints). The cost function used for TOM-P is a weighted function of the mission total obtained weight, mission time, distance travelled, and number of tasks.

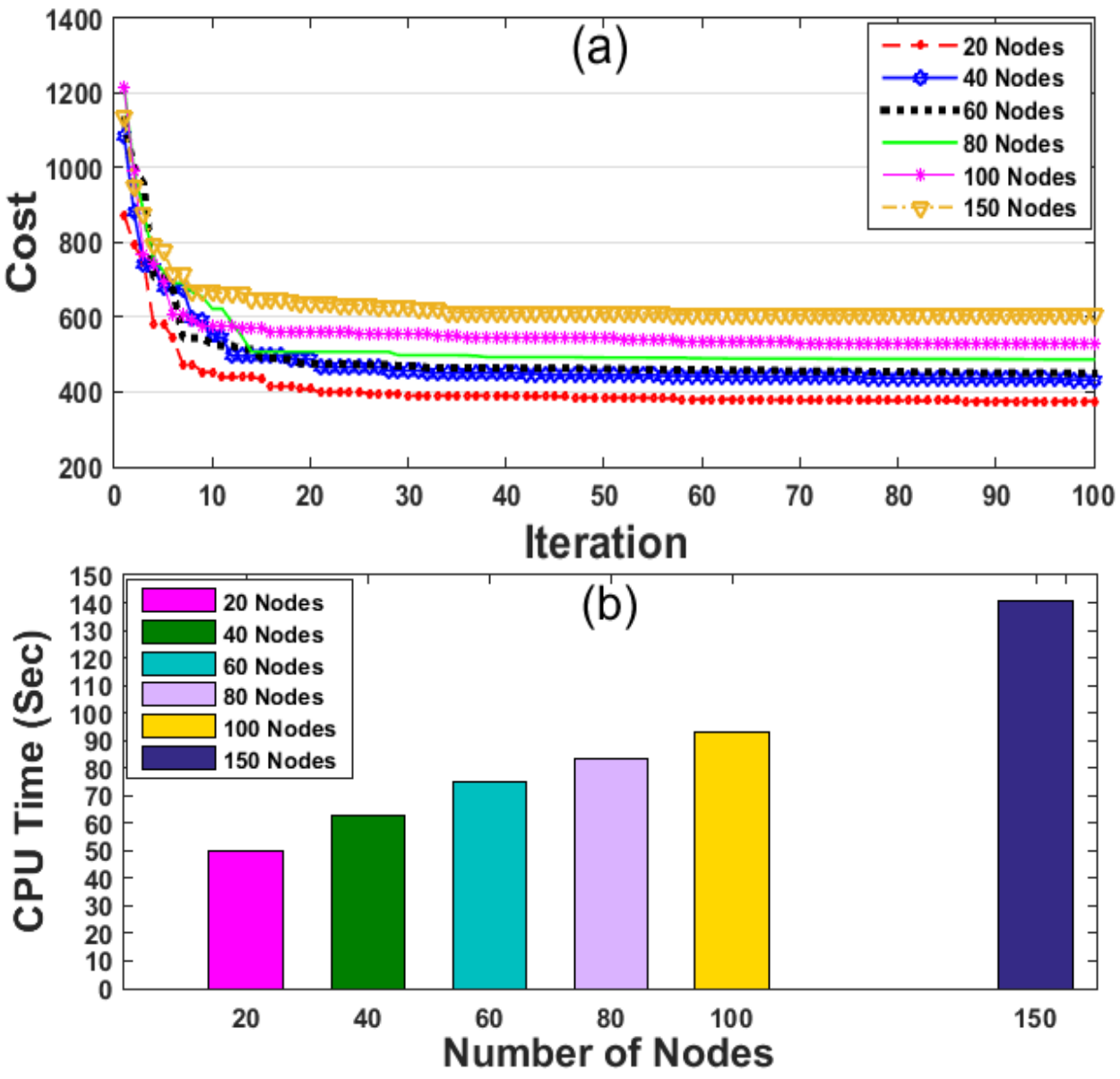

**Fig.12. (a)** The mission cost over 100 iterations for different graph complexities; **(b)** CPU time variation to graph complexity.

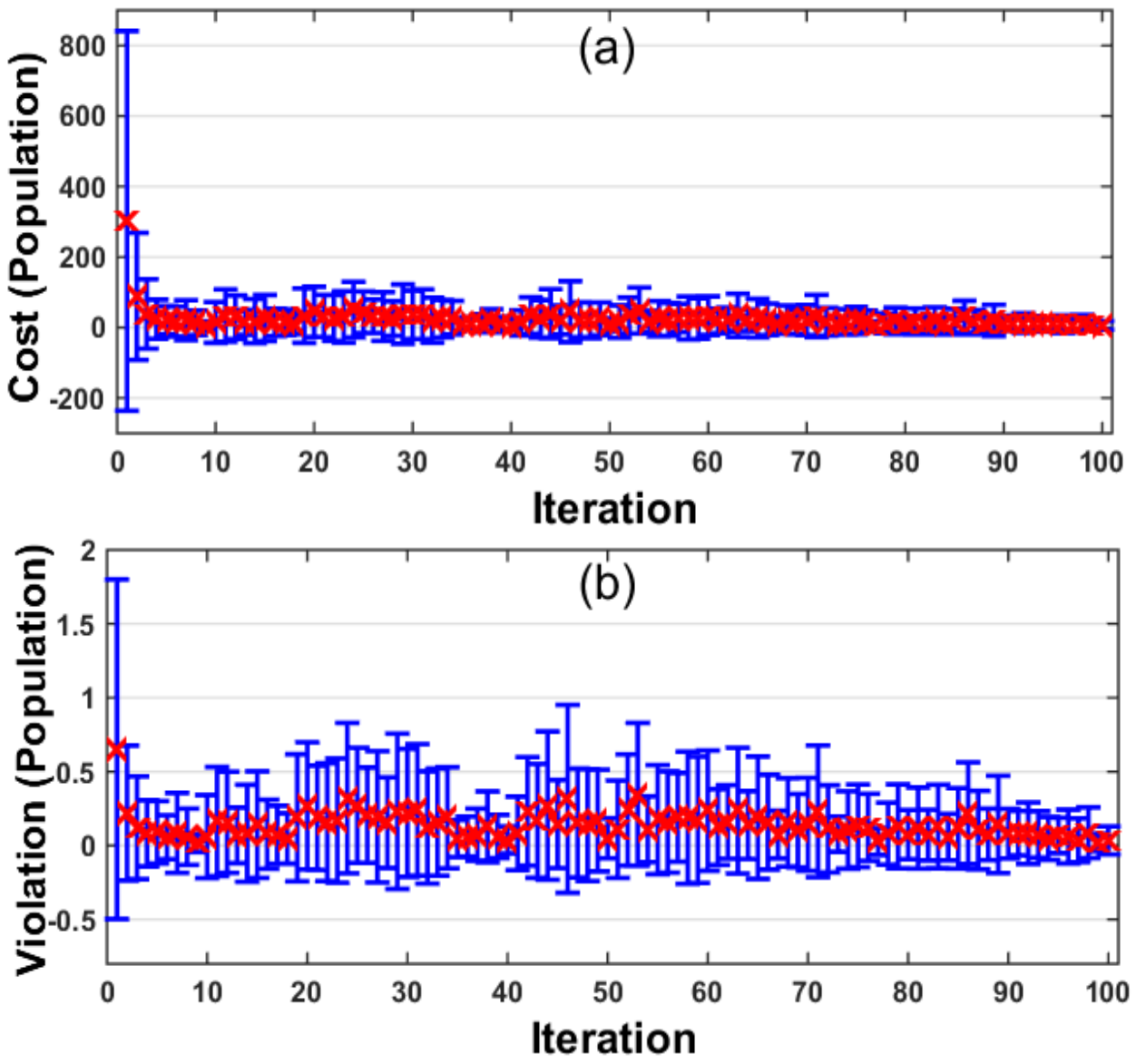

**Fig.13. (a)** The average cost variations over 100 iterations for a graph with 80 nodes; **(b)** Iterative violation of the solutions to time restrictions.

As presented in *Fig*.12(a), the algorithm accurately manages the complexity growth as the cost value decrements iteratively regardless of size and topology complexity of the network. Considering *Fig*.12(b), the computational time increments linearly with growth of the graph nodes and the CPU time for all samples remain within the bounds of a suitable real-time solution. The cost and CPU-Time variations in *Fig*.12(a,b) declares that the performance of the model is almost independent of both size and complexity of the graph, whereas this is recognized as a problematic issue in many of the outlined approaches. The *Fig*.13 indicates the iterative variation of cost and violation of the solutions in satisfying specified objectives and constraints.

It is further inferred from *Fig*.13(a,b), the algorithm efficiently tends to minimise the cost over 100 iterations and the cost variation range is reduced iteratively which means the population effectively converges to the optimum solution with the minimum cost. The violation diagram shows how the generated solutions respects the defined restriction (restriction of mission time to available time and feasibility criteria). Tracking the variation of red crosses in the middle of the error bar graphs that corresponds to mean cost and mean violation, declares that the algorithm accurately enforces the solutions to approach the least cost and manage the solutions to eliminate the violation as the range of violation variations converges to zero within 100 iterations.

The 200 Monte Carlo simulation runs are carried out and performed by *Fig*.14 to show the stability of the applied method in a quantitative manner, in which the variation ranges of performance metrics of total obtained weight, completed tasks, total travelled time and distance, CPU time, and mission cost are presented. By analyzing the result of the Monte Carlo simulations, it is possible to be confident in the robustness and efficiency of the method in dealing with random transformation of the graph size and topology. For all Monte Carlo runs, the quantity of waypoint is set to be changed with a uniform distribution between 20 to 50 nodes and the network topology also transforms randomly with Gaussian distribution on the problem search space. The time threshold is set on $3.6\times10^{4}$(*sec*). According to results presented in *Fig*.14, the average travelled distance is approximately drawn around 80 *km* and within this distance nearly 8 to 16 number of tasks is completed with the average weight about 30 to 40 worth.*Fig*.14 shows consistency of the generated solutions by ACO in its distribution dealing with problem space deformation.

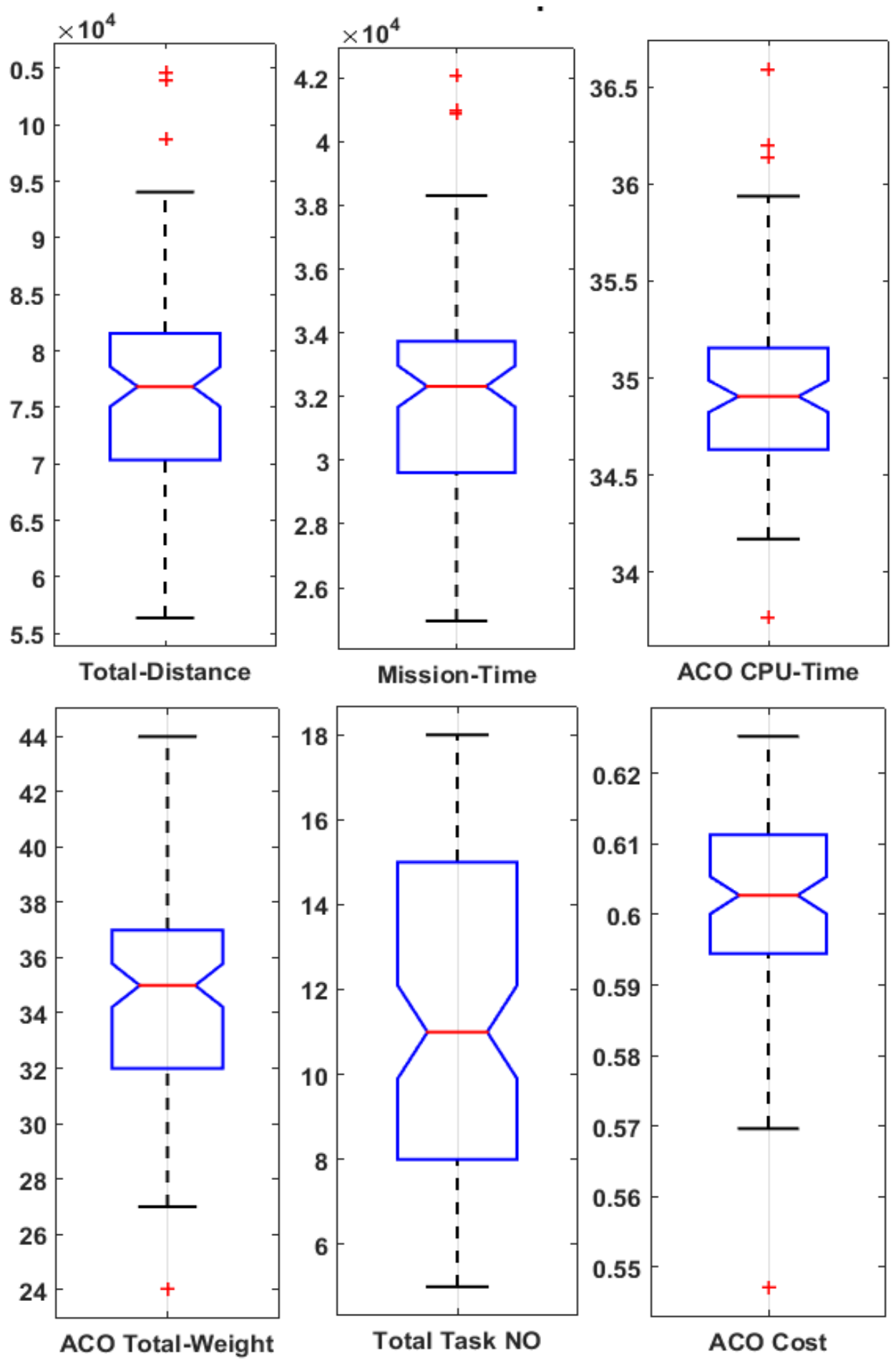

**Fig.14.** Statistical analysis of ACO in terms of satisfying given performance metrics in 200 Monte Carlo simulations

### 4.3 Architecture's Performance on Scheduling and Time Management for Reliable and Efficient Operation

At the top level, the TOM-P uses offline map in the large scale to find the best fitted task sequence to the predefined available time in order to maximize the mission productivity. On the other hand, the autonomous vehicle needs to perceive the new information of the dynamic environment, comprehend the relation between perceived information and meaning of different raised situation to be able to handle raised unexpected changes of the terrain. Hence, dynamic changes of the operation field that handled by LOP-P module in smaller scale.

Quality of the generated local path also affects the optimality and productivity of the TOM-P generated mission plan, in which the mission re-planning is performed to rearrange the tasks order or cut off some of them in cases that available time is expended more than expectation for incorporating raised changes. Considering updated available time and environmental changes, the TOM-P is dynamically re-compute the mission plan until the vehicle reaches to the target point.

The particular objective of the entire architecture is to guarantee a reliable mission and accurate arrangement of the prioritised tasks so that the vehicle is guided toward the destination while takes the maximum use of available time and reach the destination before it runs out of time. The architecture capability in time management and guarantying on-time termination of the mission is investigated with 10 individual experiments as is shown in *Fig*.15. To establish an accurate and stringent concordance among different components of the architecture and real-time implementation of the model some other performance indexes should be highlighted additional to those which addressed in two previous sections. The first critical factor for both TOM-P and LOP-P operation is having a short computational time to provide a concurrent synchronization between modules and balancing the correlation between their operations. A fast operation for each component of the main architecture prevent any of them from dropping behind the process of the other components, since appearance of such a delay interrupts the routine flow and concurrency of the entire system. Another significant performance indicator is compatibility of the value of the local path time ($T_{\wp}$) and the expected time ($T_{\varepsilon}$) in simultaneous operation of the LOP-P module in context of the main architecture; hence, a big difference between values of $T_{\wp}$ and $T_{\varepsilon}$ is not acceptable. This issue is critical for recognizing the requisition for re-routing and re-scheduling the tasks and again influences the balance and synchronism of the whole system. To validate the performance of the introduced architecture in a quantitative feature, the concurrent and collaborative procedure of engaged modules toward satisfying addressed objectives is investigated through multiple experiments with the initial condition analogous to real underwater mission scenarios indicated by Table.2 and *Fig*.15. Accordingly, the operation network is initialized with 50 waypoints and 1600 edges, in which the start and final destination waypoints are assigned with nodes 1 and 50, respectively. A fixed collection of prioritized tasks is specified and randomly assigned to some of the edges. The edges that doesn't assigned with a task are weighted with 1. The location of nodes are randomized according to $\{P^{i}_{x,y}\sim\mathbf{U}(0,10000)$ and $P^{i}_{z}\sim\mathbf{U}(0,100)\}$. The $T_{Total}$ is initialized with 10800(*sec*) for all experiments. The terrain is modelled as a realistic underwater environment, in which time varying ocean current and randomly generated uncertain static/moving obstacles are included.

**Table.1.** Notation table representing the elements of Table.2 (to prevent confusion)

| | |
|---|---|
| **PP** | Number of recall to LOP-P |
| **$E_G$** | Edges of the graph selected for path planning |
| **$\wp\nabla$** | Total violation for the local path |
| **$\wp C$** | Path Cost |
| **$T_\wp$** | Time for the generated path |
| **$T_\varepsilon$** | The expected time for travelling a path |
| **$FM_{re}$** | Mission re-planning flag |
| **$N_{Tsk}$** | Number of tasks to be completed in a mission |
| **$W_M$** | Total weight collected in a mission |
| **$T_M$** | Total estimated mission time |
| **$T_{Total}$** | Total available time left for mission |
| **MC** | Mission Cost |
| **$T_{CPU}$** | Computational time |
| **$M_{SEQ}$** | Planned mission-in a sequence of waypoints (Tasks) |
| **Valid** | Feasibility and Valid of the $M_{SEQ}$ {Valid:1, not Valid:0} |

**Table.2.**Collaborative and concurrent process of the TOM-P and LOP-P modules in the proposed architecture.

Table.2(a). Experiment-1

| | | | | | | | | |
|---|---|---|---|---|---|---|---|---|
| **$M_{SEQ}$-1:**{ 1-6-3-20-12-17-42-29-50 } | | | | | | | | |
| $N_{Tsk}$ | $W_M$ | Valid | MC | $T_{CPU}$ | $T_M$ | | $T_{Total}$ | |
| 8 | 27 | 1 | 0.011 | 20.8 | 10458 | | 10800 | |
| PP | $E_G$ | $\wp\nabla$ | $\wp C$ | $T_{CPU}$ | $T_\wp$ | $T_\varepsilon$ | $T_{Total}$ | $FM_{re}$ |
| 1 | 1-6 | 0.0000 | 0.149 | 43.8 | 2042.0 | 2210.0 | 8758.0 | 0 |
| 2 | 6-3 | 0.000 | 0.602 | 38.0 | 2132.7 | 2168.0 | 6625.4 | 0 |
| 3 | 3-20 | 0.0030 | 0.105 | 46.3 | 373.70 | 528.30 | 6251.6 | 0 |
| 4 | 20-12 | 0.0000 | 0.540 | 44.3 | 1806.4 | 1829.7 | 4445.2 | 0 |
| 5 | 12-17 | 0.0002 | 0.609 | 39.8 | 514.00 | 333.40 | 3931.2 | 0 |
| **$M_{SEQ}$**-2:{ 17-29-7-27-8-10-50 } | | | | | | | | |
| $N_{Tsk}$ | $W_M$ | Valid | MC | $T_{CPU}$ | $T_M$ | | $T_{Total}$ | |
| 6 | 25 | 1 | 0.030 | 18.3 | 3802 | | 3931 | |
| PP | $E_G$ | $\wp\nabla$ | $\wp C$ | $T_{CPU}$ | $T_\wp$ | $T_\varepsilon$ | $T_{Total}$ | $FM_{re}$ |
| 1 | 17-29 | 0.0000 | 0.230 | 47.1 | 773.50 | 835.60 | 3157.6 | 0 |
| 2 | 29-7 | 0.0011 | 0.321 | 43.3 | 1071.0 | 1000.0 | 2085.8 | 1 |
| **$M_{SEQ}$** --3:{ 7-15-50 } | | | | | | | | |
| $N_{Tsk}$ | $W_M$ | Valid | MC | $T_{CPU}$ | $T_M$ | | $T_{Total}$ | |
| 2 | 24 | 1 | 0.540 | 16.8 | 2022 | | 2086 | |
| PP | $E_G$ | $\wp\nabla$ | $\wp C$ | $T_{CPU}$ | $T_\wp$ | $T_\varepsilon$ | $T_{Total}$ | $FM_{re}$ |
| 1 | 7-15 | 0.0000 | 0.207 | 32.8 | 705.03 | 816.60 | 1380.8 | 0 |
| 2 | 15-50 | 0.0000 | 0.304 | 36.1 | 1021.1 | 1076.0 | 359.70 | done |

Table.2(b). Experiment-2

| | | | | | | | | |
|---|---|---|---|---|---|---|---|---|
| **$M_{SEQ}$**-1:{ 1-39-7-16-48-33-40-38-50 } | | | | | | | | |
| $N_{Tsk}$ | $W_M$ | Valid | MC | $T_{CPU}$ | $T_M$ | | $T_{Total}$ | |
| 8 | 42 | 1 | 0.048 | 17.3 | 10262 | | 10800 | |
| PP | $E_G$ | $\wp\nabla$ | $\wp C$ | $T_{CPU}$ | $T_\wp$ | $T_\varepsilon$ | $T_{Total}$ | $FM_{re}$ |
| 1 | 1-39 | 0.0000 | 0.226 | 47.3 | 2333.1 | 2535.3 | 8466.7 | 0 |
| 2 | 39-7 | 0.0043 | 0.701 | 39.8 | 755.8 | 666.6 | 7710.8 | 1 |
| **$M_{SEQ}$**-2:{ 4-71-14-12-36-48-22-15-47-40-50 } | | | | | | | | |
| $N_{Tsk}$ | $W_M$ | Valid | MC | $T_{CPU}$ | $T_M$ | | $T_{Total}$ | |
| 10 | 38 | 1 | 0.048 | 17.3 | 7805 | | 7710.8 | |
| PP | $E_G$ | $\wp\nabla$ | $\wp C$ | $T_{CPU}$ | $T_\wp$ | $T_\varepsilon$ | $T_{Total}$ | $FM_{re}$ |
| 1 | 7-41 | 0.0000 | 0.146 | 42.4 | 501.4 | 508.3 | 7209.4 | 0 |
| 2 | 41-14 | 0.0012 | 0.3170 | 40.0 | 1078 | 1179 | 6131.4 | 0 |
| 3 | 14-12 | 0.0000 | 0.226 | 42.3 | 756.1 | 686.6 | 5375.3 | 0 |
| **$M_{SEQ}$**-3:{ 12-4-39-44-30-28-11-50 } | | | | | | | | |
| $N_{Tsk}$ | $W_M$ | Valid | MC | $T_{CPU}$ | $T_M$ | | $T_{Total}$ | |
| 7 | 36 | 1 | 0.024 | 21.9 | 5211 | | 5375.3 | |
| PP | $E_G$ | $\wp\nabla$ | $\wp C$ | $T_{CPU}$ | $T_\wp$ | $T_\varepsilon$ | $T_{Total}$ | $FM_{re}$ |
| 1 | 12-4 | 0.0000 | 0.279 | 40.9 | 1140.6 | 528.3 | 4234.7 | 1 |
| **$M_{SEQ}$**-4:{ 4-41-12-44-11-50 } | | | | | | | | |
| $N_{Tsk}$ | $W_M$ | Valid | MC | $T_{CPU}$ | $T_M$ | | $T_{Total}$ | |
| 5 | 34 | 1 | 0.040 | 19.8 | 3998 | | 4234.7 | |
| PP | $E_G$ | $\wp\nabla$ | $\wp C$ | $T_{CPU}$ | $T_\wp$ | $T_\varepsilon$ | $T_{Total}$ | $FM_{re}$ |
| 1 | 4-41 | 0.0006 | 0.202 | 37.4 | 760.02 | 696.8 | 3474.7 | 1 |
| **$M_{SEQ}$**-5:{ 41-46-3-44-17-50 } | | | | | | | | |
| $N_{Tsk}$ | $W_M$ | Valid | MC | $T_{CPU}$ | $T_M$ | | $T_{Total}$ | |
| 5 | 36 | 1 | 0.029 | 18.4 | 3468 | | 3474.7 | |
| PP | $E_G$ | $\wp\nabla$ | $\wp C$ | $T_{CPU}$ | $T_\wp$ | $T_\varepsilon$ | $T_{Total}$ | $FM_{re}$ |
| 1 | 41-46 | 0.0000 | 0.201 | 40.6 | 674.5 | 820.6 | 2800.2 | 0 |
| 2 | 46-3 | 0.0028 | 0.207 | 44.1 | 682.8 | 647.8 | 2117.3 | 1 |
| **$M_{SEQ}$**-6:{ 3-29-30-42-50 } | | | | | | | | |
| $N_{Tsk}$ | $W_M$ | Valid | MC | $T_{CPU}$ | $T_M$ | | $T_{Total}$ | |
| 4 | 38 | 1 | 0.078 | 20.7 | 2008 | | 2117.3 | |
| PP | $E_G$ | $\wp\nabla$ | $\wp C$ | $T_{CPU}$ | $T_\wp$ | $T_\varepsilon$ | $T_{Total}$ | $FM_{re}$ |
| 1 | 3-29 | 0.0032 | 0.160 | 39.8 | 567.2 | 857.6 | 1550.1 | 0 |
| 2 | 29-30 | 0.0000 | 0.146 | 43.4 | 495.5 | 334.8 | 1054.5 | 1 |
| **$M_{SEQ}$**-7:{ 30-42-50 } | | | | | | | | |
| $N_{Tsk}$ | $W_M$ | Valid | MC | $T_{CPU}$ | $T_M$ | | $T_{Total}$ | |
| 2 | 14 | 1 | 0.477 | 21.8 | 1051 | | 1054.5 | |
| PP | $E_G$ | $\wp\nabla$ | $\wp C$ | $T_{CPU}$ | $T_\wp$ | $T_\varepsilon$ | $T_{Total}$ | $FM_{re}$ |
| 1 | 30-42 | 0.0000 | 0.142 | 39.7 | 479.04 | 482.1 | 575.4 | 0 |
| 2 | 42-50 | 0.0002 | 0.137 | 40.1 | 563.7 | 569.3 | 11.7 | done |

Table.2(c). Experiment-3

| | | | | | | | | |
|---|---|---|---|---|---|---|---|---|
| **$M_{SEQ}$**-1:{ 1-2-29-35-30-50 } | | | | | | | | |
| $N_{Tsk}$ | $W_M$ | Valid | MC | $T_{CPU}$ | $T_M$ | | $T_{Total}$ | |
| 5 | 26 | 1 | 0.420 | 23.3 | 10248 | | 10800 | |
| PP | $E_G$ | $\wp\nabla$ | $\wp C$ | $T_{CPU}$ | $T_\wp$ | $T_\varepsilon$ | $T_{Total}$ | $FM_{re}$ |
| 1 | 1-2 | 0.0000 | 0.736 | 36.7 | 2539.2 | 2535.3 | 8260.8 | 1 |
| **$M_{SEQ}$**-2:{ 2-23-25-16-4-46-50} | | | | | | | | |
| $N_{Tsk}$ | $W_M$ | Valid | MC | $T_{CPU}$ | $T_M$ | | $T_{Total}$ | |
| 6 | 25 | 1 | 0.440 | 23.1 | 7844 | | 8260.8 | |
| PP | $E_G$ | $\wp\nabla$ | $\wp C$ | $T_{CPU}$ | $T_\wp$ | $T_\varepsilon$ | $T_{Total}$ | $FM_{re}$ |
| 1 | 2-23 | 0.0000 | 0.730 | 40.1 | 2456.5 | 2526.4 | 5804.3 | 0 |
| 2 | 23-25 | 0.0009 | 0.150 | 39.6 | 504.9 | 389.3 | 5299.3 | 1 |
| **$M_{SEQ}$**-3:{ 25-5-35-43-18-9-50 } | | | | | | | | |
| $N_{Tsk}$ | $W_M$ | Valid | MC | $T_{CPU}$ | $T_M$ | | $T_{Total}$ | |
| 6 | 28 | 1 | 0.250 | 20.9 | 5036 | | 5299.3 | |
| PP | $E_G$ | $\wp\nabla$ | $\wp C$ | $T_{CPU}$ | $T_\wp$ | $T_\varepsilon$ | $T_{Total}$ | $FM_{re}$ |
| 1 | 25-5 | 0.0000 | 0.429 | 40.5 | 1444.3 | 1534.3 | 3855 | 0 |
| 2 | 5-35 | 0.0000 | 0.151 | 41.3 | 514.3 | 839.6 | 3340.6 | 0 |
| 3 | 35-43 | 0.0071 | 0.318 | 39.0 | 1084.4 | 1000.8 | 2256.2 | 1 |
| **$M_{SEQ}$**-4:{ 43-7-9-50 } | | | | | | | | |
| $N_{Tsk}$ | $W_M$ | Valid | MC | $T_{CPU}$ | $T_M$ | | $T_{Total}$ | |
| 3 | 13 | 1 | 0.850 | 21.1 | 2196 | | 2256.2 | |
| PP | $E_G$ | $\wp\nabla$ | $\wp C$ | $T_{CPU}$ | $T_\wp$ | $T_\varepsilon$ | $T_{Total}$ | $FM_{re}$ |
| 1 | 43-7 | 0.0000 | 1.831 | 42.1 | 529.9 | 496.3 | 1726.3 | 1 |
| **$M_{SEQ}$**-5:{ 7-50 } | | | | | | | | |
| $N_{Tsk}$ | $W_M$ | Valid | MC | $T_{CPU}$ | $T_M$ | | $T_{Total}$ | |
| 1 | 8 | 1 | 0.980 | 19.8 | 870 | | 1726.3 | |
| PP | $E_G$ | $\wp\nabla$ | $\wp C$ | $T_{CPU}$ | $T_\wp$ | $T_\varepsilon$ | $T_{Total}$ | $FM_{re}$ |
| 1 | 7-50 | 0.0000 | 0.223 | 39.8 | 749.9 | 870.8 | 976.3 | done |

Table.2(d). Experiment-4

| | | | | | | | | |
|---|---|---|---|---|---|---|---|---|
| **$M_{SEQ}$**-1:{ 1-25-38-44-9-11-31-27-12-50 } | | | | | | | | |
| $N_{Tsk}$ | $W_M$ | Valid | MC | $T_{CPU}$ | $T_M$ | | $T_{Total}$ | |
| 9 | 43 | 1 | 0.430 | 28.3 | 10260 | | 10800 | |
| PP | $E_G$ | $\wp\nabla$ | $\wp C$ | $T_{CPU}$ | $T_\wp$ | $T_\varepsilon$ | $T_{Total}$ | $FM_{re}$ |
| 1 | 1-25 | 0.0000 | 0.206 | 48.3 | 688.3 | 668.4 | 10112 | 1 |
| **$M_{SEQ}$**-2:{ 25-15-44-22-20-48-12-50 } | | | | | | | | |
| $N_{Tsk}$ | $W_M$ | Valid | MC | $T_{CPU}$ | $T_M$ | | $T_{Total}$ | |
| 7 | 33 | 1 | 0.420 | 23.2 | 9870 | | 10112 | |
| PP | $E_G$ | $\wp\nabla$ | $\wp C$ | $T_{CPU}$ | $T_\wp$ | $T_\varepsilon$ | $T_{Total}$ | $FM_{re}$ |
| 1 | 25-15 | 0.0001 | 0.228 | 44.1 | 765.89 | 818.6 | 9345.8 | 0 |
| 2 | 15-44 | 0.0000 | 0.419 | 34.5 | 1400.2 | 1533.3 | 7945.5 | 0 |
| 3 | 44-22 | 0.0000 | 0.227 | 39.6 | 765 | 864.6 | 7180.5 | 0 |
| 4 | 22-20 | 0.0000 | 0.143 | 39.3 | 477.9 | 539.3 | 6702.5 | 0 |
| 5 | 20-48 | 0.0007 | 0.233 | 44.8 | 807.7 | 697.3 | 5894.8 | 1 |
| **$M_{SEQ}$**-3:{ 48-4-32-18-41-23-33-50 } | | | | | | | | |
| $N_{Tsk}$ | $W_M$ | Valid | MC | $T_{CPU}$ | $T_M$ | | $T_{Total}$ | |
| 7 | 37 | 1 | 1.040 | 26.1 | 5437 | | 5894.8 | |
| PP | $E_G$ | $\wp\nabla$ | $\wp C$ | $T_{CPU}$ | $T_\wp$ | $T_\varepsilon$ | $T_{Total}$ | $FM_{re}$ |
| 1 | 48-4 | 0.0000 | 0.537 | 41.7 | 1796.5 | 1828.7 | 4098.3 | 0 |
| 2 | 4-32 | 0.0018 | 0.107 | 40.8 | 357.2 | 334.5 | 3741 | 1 |
| **$M_{SEQ}$**-4:{ 32-9-3-31-50 } | | | | | | | | |
| $N_{Tsk}$ | $W_M$ | Valid | MC | $T_{CPU}$ | $T_M$ | | $T_{Total}$ | |
| 4 | 31 | 1 | 0.200 | 19.7 | 3018 | | 3741 | |
| PP | $E_G$ | $\wp\nabla$ | $\wp C$ | $T_{CPU}$ | $T_\wp$ | $T_\varepsilon$ | $T_{Total}$ | $FM_{re}$ |
| 1 | 32-9 | 0.0000 | 0.241 | 39.3 | 815.4 | 865.6 | 2925.6 | 0 |
| 2 | 9-3 | 0.0000 | 0.152 | 40.9 | 514.7 | 387.2 | 2410.8 | 1 |
| **$M_{SEQ}$**-5:{ 3-27-41-18-19-43-50 } | | | | | | | | |
| $N_{Tsk}$ | $W_M$ | Valid | MC | $T_{CPU}$ | $T_M$ | | $T_{Total}$ | |
| 6 | 23 | 1 | 0.950 | 25.1 | 2398 | | 2410.8 | |
| PP | $E_G$ | $\wp\nabla$ | $\wp C$ | $T_{CPU}$ | $T_\wp$ | $T_\varepsilon$ | $T_{Total}$ | $FM_{re}$ |
| 1 | 3-27 | 0.0023 | 0.238 | 40.4 | 797.1 | 684.4 | 1613.7 | 1 |
| **$M_{SEQ}$**-6:{ 27-41-46-50 } | | | | | | | | |
| $N_{Tsk}$ | $W_M$ | Valid | MC | $T_{CPU}$ | $T_M$ | | $T_{Total}$ | |
| 3 | 19 | 1 | 0.218 | 27.6 | 1412 | | 1613.7 | |
| PP | $E_G$ | $\wp\nabla$ | $\wp C$ | $T_{CPU}$ | $T_\wp$ | $T_\varepsilon$ | $T_{Total}$ | $FM_{re}$ |
| 1 | 27-41 | 0.0000 | 0.123 | 37.8 | 430.49 | 347.3 | 1183.2 | 1 |
| **$M_{SEQ}$**-7:{ 41-50 } | | | | | | | | |
| $N_{Tsk}$ | $W_M$ | Valid | MC | $T_{CPU}$ | $T_M$ | | $T_{Total}$ | |
| 1 | 7 | 1 | 1.760 | 20.0 | 674 | | 1183.2 | |
| PP | $E_G$ | $\wp\nabla$ | $\wp C$ | $T_{CPU}$ | $T_\wp$ | $T_\varepsilon$ | $T_{Total}$ | $FM_{re}$ |
| 1 | 41-50 | 0.0000 | 0.206 | 41.3 | 668.2 | 673.7 | 514.9 | done |

Table.2(g). Experiment-5

| | | | | | | | | |
|---|---|---|---|---|---|---|---|---|
| **$M_{SEQ}$**-1: { 1-27-11-7-13-21-18-50 } | | | | | | | | |
| $N_{Tsk}$ | $W_M$ | Valid | MC | $T_{CPU}$ | $T_M$ | | $T_{Total}$ | |
| 7 | 41 | 1 | 0.038 | 27.6 | 10792 | | 10800 | |
| PP | $E_G$ | $\wp\nabla$ | $\wp C$ | $T_{CPU}$ | $T_\wp$ | $T_\varepsilon$ | $T_{Total}$ | $FM_{re}$ |
| 1 | 1-27 | 0.0003 | 0.430 | 33.6 | 1701.3 | 1703.1 | 9098.7 | 0 |
| 2 | 27-11 | 0.0000 | 0.360 | 37.4 | 1347.1 | 1351.3 | 7751.6 | 0 |
| 3 | 11-7 | 0.0000 | 0.580 | 41.6 | 478.9 | 480.2 | 7272.6 | 0 |
| 4 | 7-13 | 0.0014 | 0.230 | 42.7 | 1228.1 | 1231 | 6044.5 | 0 |
| 5 | 13-21 | 0.0000 | 0.470 | 44.6 | 1690.6 | 1693.2 | 4353.9 | 0 |
| 6 | 21-18 | 0.0008 | 0.380 | 39.0 | 1396.5 | 1401.3 | 2957.5 | 0 |
| 7 | 18-50 | 0.0000 | 0.170 | 39.7 | 2925.4 | 2931.2 | 31.1 | done |

| Table.2(e). Experiment-6 | | | | | | | | |
|---|---|---|---|---|---|---|---|---|
| **$M_{SEQ}$**-1:{ 1-24-7-25-42-11-26-44-50 } | | | | | | | | |
| $N_{Tsk}$ | $W_M$ | Valid | MC | $T_{CPU}$ | $T_M$ | | $T_{Total}$ | |
| 8 | 38 | 1 | 0.430 | 23.1 | 10670 | | 10800 | |
| PP | $E_G$ | $\wp\nabla$ | $\wp C$ | $T_{CPU}$ | $T_\wp$ | $T_\varepsilon$ | $T_{Total}$ | $FM_{re}$ |
| 1 | 1-24 | 0.0000 | 0.450 | 41.6 | 1532.7 | 1666.7 | 9267.3 | 0 |
| 2 | 24-7 | 0.0005 | 0.510 | 40.3 | 1702.3 | 1872.7 | 7565 | 0 |
| 3 | 7-25 | 0.0000 | 0.460 | 36.4 | 1702.1 | 1673.1 | 5862.8 | 1 |
| **$M_{SEQ}$**-2:{ 25-36-26-27-33-5-50 } | | | | | | | | |
| $N_{Tsk}$ | $W_M$ | Valid | MC | $T_{CPU}$ | $T_M$ | | $T_{Total}$ | |
| 6 | 33 | 1 | 0.320 | 20.9 | 5831 | | 5862.8 | |
| PP | $E_G$ | $\wp\nabla$ | $\wp C$ | $T_{CPU}$ | $T_\wp$ | $T_\varepsilon$ | $T_{Total}$ | $FM_{re}$ |
| 1 | 25-36 | 0.0000 | 0.115 | 37.8 | 467.2 | 535.3 | 5395.6 | 0 |
| 2 | 36-26 | 0.0009 | 0.311 | 43.7 | 1053.2 | 1210 | 4342.4 | 0 |
| 3 | 26-27 | 0.0000 | 0.369 | 39.1 | 1306.2 | 1333.4 | 3036.3 | 0 |
| 4 | 27-33 | 0.0014 | 0.232 | 39.7 | 778.7 | 668.3 | 2257.6 | 1 |
| **$M_{SEQ}$**-3:{ 33-50 } | | | | | | | | |
| $N_{Tsk}$ | $W_M$ | Valid | MC | $T_{CPU}$ | $T_M$ | | $T_{Total}$ | |
| 1 | 14 | 1 | 0.610 | 19.8 | 1828 | | 2257.6 | |
| PP | $E_G$ | $\wp\nabla$ | $\wp C$ | $T_{CPU}$ | $T_\wp$ | $T_\varepsilon$ | $T_{Total}$ | $FM_{re}$ |
| 1 | 33-50 | 0.0008 | 0.511 | 40.6 | 705.4 | 1827.7 | 552.1 | done |

| Table.2(k). Experiment-7 | | | | | | | | |
|---|---|---|---|---|---|---|---|---|
| **$M_{SEQ}$**-1:{ 1-23-36-28-3-15-47-50 } | | | | | | | | |
| $N_{Tsk}$ | $W_M$ | Valid | MC | $T_{CPU}$ | $T_M$ | | $T_{Total}$ | |
| 7 | 38 | 1 | 0.420 | 23.6 | 10388 | | 10800 | |
| PP | $E_G$ | $\wp\nabla$ | $\wp C$ | $T_{CPU}$ | $T_\wp$ | $T_\varepsilon$ | $T_{Total}$ | $FM_{re}$ |
| 1 | 1-23 | 0.0000 | 0.328 | 36.7 | 2476 | 2514.3 | 8324 | 0 |
| 2 | 2336 | 0.0000 | 0.287 | 44.1 | 965 | 1183 | 7358.9 | 0 |
| 3 | 36-28 | 0.0008 | 0.743 | 37.8 | 486.7 | 333.8 | 6872.2 | 1 |
| **$M_{SEQ}$**-2:{ 28-18-9-3-24-7-29-50 } | | | | | | | | |
| $N_{Tsk}$ | $W_M$ | Valid | MC | $T_{CPU}$ | $T_M$ | | $T_{Total}$ | |
| 7 | 35 | 1 | 0.340 | 26.1 | 6531 | | 6872.2 | |
| PP | $E_G$ | $\wp\nabla$ | $\wp C$ | $T_{CPU}$ | $T_\wp$ | $T_\varepsilon$ | $T_{Total}$ | $FM_{re}$ |
| 1 | 28-18 | 0.0001 | 0.109 | 39.6 | 519.6 | 376.3 | 6352.5 | 1 |
| **$M_{SEQ}$**-3:{ 18-5-25-35-8-50 } | | | | | | | | |
| $N_{Tsk}$ | $W_M$ | Valid | MC | $T_{CPU}$ | $T_M$ | | $T_{Total}$ | |
| 5 | 23 | 1 | 0.270 | 23.4 | 6112 | | 6352.5 | |
| PP | $E_G$ | $\wp\nabla$ | $\wp C$ | $T_{CPU}$ | $T_\wp$ | $T_\varepsilon$ | $T_{Total}$ | $FM_{re}$ |
| 1 | 18-5 | 0.0000 | 0.120 | 44.5 | 1406 | 1492.3 | 4946.6 | 0 |
| 2 | 5-25 | 0.0000 | 0.101 | 43.6 | 344.6 | 536.3 | 4601.9 | 0 |
| 3 | 25-35 | 0.0040 | 0.360 | 44.3 | 1272.6 | 1495.3 | 3329.3 | 0 |
| 4 | 35-8 | 0.0000 | 0.416 | 44.9 | 1068.3 | 1000.3 | 2260.9 | 1 |
| **$M_{SEQ}$**-4:{ 8-40-7-50 } | | | | | | | | |
| $N_{Tsk}$ | $W_M$ | Valid | MC | $T_{CPU}$ | $T_M$ | | $T_{Total}$ | |
| 3 | 13 | 1 | 1.660 | 19.6 | 1978 | | 2269.9 | |
| PP | $E_G$ | $\wp\nabla$ | $\wp C$ | $T_{CPU}$ | $T_\wp$ | $T_\varepsilon$ | $T_{Total}$ | $FM_{re}$ |
| 1 | 8-40 | 0.0006 | 0.319 | 38.7 | 1106.3 | 1000.8 | 1154.7 | 1 |
| **$M_{SEQ}$**-5:{ 40-50 } | | | | | | | | |
| $N_{Tsk}$ | $W_M$ | Valid | MC | $T_{CPU}$ | $T_M$ | | $T_{Total}$ | |
| 1 | 7 | 1 | 1.830 | 20.6 | 850 | | 1154.7 | |
| PP | $E_G$ | $\wp\nabla$ | $\wp C$ | $T_{CPU}$ | $T_\wp$ | $T_\varepsilon$ | $T_{Total}$ | $FM_{re}$ |
| 1 | 40-50 | 0.0001 | 0.233 | 40.2 | 781.8 | 848.6 | 372.7 | done |

| **$M_{SEQ}$**-2(m). Experiment-8 | | | | | | | | |
|---|---|---|---|---|---|---|---|---|
| **$M_{SEQ}$**-1:{ 1-30-22-19-36-44-50 } | | | | | | | | |
| $N_{Tsk}$ | $W_M$ | Valid | MC | $T_{CPU}$ | $T_M$ | | $T_{Total}$ | |
| 6 | 29 | 1 | 0.450 | 24.1 | 10473 | | 10800 | |
| PP | $E_G$ | $\wp\nabla$ | $\wp C$ | $T_{CPU}$ | $T_\wp$ | $T_\varepsilon$ | $T_{Total}$ | $FM_{re}$ |
| 1 | 1-30 | 0.0000 | 0.366 | 39.3 | 1236.1 | 1492.3 | 9563.9 | 0 |
| 2 | 30-22 | 0.0062 | 0.413 | 41.2 | 1387.5 | 1529.3 | 8176 | 0 |
| 3 | 22-19 | 0.0000 | 0.123 | 42.1 | 412.4 | 522.3 | 7764 | 0 |
| 4 | 19-36 | 0.0009 | 0.720 | 42.5 | 2466.6 | 2333.4 | 5297.4 | 1 |
| **$M_{SEQ}$**-2:{ 36-48-22-34-50 } | | | | | | | | |
| $N_{Tsk}$ | $W_M$ | Valid | MC | $T_{CPU}$ | $T_M$ | | $T_{Total}$ | |
| 4 | 27 | 1 | 0.210 | 21.3 | 4890 | | 5297.4 | |
| PP | $E_G$ | $\wp\nabla$ | $\wp C$ | $T_{CPU}$ | $T_\wp$ | $T_\varepsilon$ | $T_{Total}$ | $FM_{re}$ |
| 1 | 36-48 | 0.0000 | 0.542 | 44.0 | 1849.5 | 1844.7 | 3447.9 | 1 |
| **$M_{SEQ}$**-3:{ 48-25-13-44-8-32-50 } | | | | | | | | |
| $N_{Tsk}$ | $W_M$ | Valid | MC | $T_{CPU}$ | $T_M$ | | $T_{Total}$ | |
| 6 | 19 | 1 | 0.930 | 26.3 | 3206 | | 3447.9 | |
| PP | $E_G$ | $\wp\nabla$ | $\wp C$ | $T_{CPU}$ | $T_\wp$ | $T_\varepsilon$ | $T_{Total}$ | $FM_{re}$ |
| 1 | 48-25 | 0.0001 | 0.106 | 39.8 | 368 | 515.8 | 3079.8 | 0 |
| 2 | 25-13 | 0.0000 | 0.224 | 39.1 | 768.2 | 835.6 | 2311.5 | 0 |
| 3 | 13-44 | 0.0000 | 0.321 | 41.3 | 1071.5 | 1000.4 | 1240 | 1 |
| **$M_{SEQ}$**-4:{ 44-32-8-50 } | | | | | | | | |
| $N_{Tsk}$ | $W_M$ | Valid | MC | $T_{CPU}$ | $T_M$ | | $T_{Total}$ | |
| 3 | 18 | 1 | 0.860 | 20.9 | 987 | | 1240 | |
| PP | $E_G$ | $\wp\nabla$ | $\wp C$ | $T_{CPU}$ | $T_\wp$ | $T_\varepsilon$ | $T_{Total}$ | $FM_{re}$ |
| 1 | 44-32 | 0.0008 | 0.103 | 43.0 | 344.2 | 323.8 | 895.7 | 1 |
| **$M_{SEQ}$**-5:{ 32-6-50 } | | | | | | | | |
| $N_{Tsk}$ | $W_M$ | Valid | MC | $T_{CPU}$ | $T_M$ | | $T_{Total}$ | |
| 2 | 12 | 1 | 1.280 | 21.1 | 873 | | 895.7 | |
| PP | $E_G$ | $\wp\nabla$ | $\wp C$ | $T_{CPU}$ | $T_\wp$ | $T_\varepsilon$ | $T_{Total}$ | $FM_{re}$ |
| 1 | 92-6 | 0.0000 | 0.128 | 43.6 | 345 | 398.3 | 550.7 | 0 |
| 2 | 6-50 | 0.0010 | 0.430 | 45.6 | 453.5 | 471.8 | 97.1 | done |

| Table.2(f). Experiment-9 | | | | | | | | |
|---|---|---|---|---|---|---|---|---|
| **$M_{SEQ}$**-1:{ 1-3-7-11-37-35-36-31-29-5-50 } | | | | | | | | |
| $N_{Tsk}$ | $W_M$ | Valid | MC | $T_{CPU}$ | $T_M$ | | $T_{Total}$ | |
| 10 | 48 | 1 | 0.450 | 20.3 | 10260 | | 10800 | |
| PP | $E_G$ | $\wp\nabla$ | $\wp C$ | $T_{CPU}$ | $T_\wp$ | $T_\varepsilon$ | $T_{Total}$ | $FM_{re}$ |
| 1 | 1-3 | 0.0000 | 0.225 | 40.3 | 1848.7 | 1873.7 | 8951.3 | 0 |
| 2 | 3-7 | 0.0000 | 0.547 | 44.2 | 755.2 | 682.6 | 8196 | 1 |
| **$M_{SEQ}$**-2:{ 7-41-1-19-38-39-27-50 } | | | | | | | | |
| $N_{Tsk}$ | $W_M$ | Valid | MC | $T_{CPU}$ | $T_M$ | | $T_{Total}$ | |
| 7 | 35 | 1 | 0.560 | 22.6 | 7890 | | 8196 | |
| PP | $E_G$ | $\wp\nabla$ | $\wp C$ | $T_{CPU}$ | $T_\wp$ | $T_\varepsilon$ | $T_{Total}$ | $FM_{re}$ |
| 1 | 7-41 | 0.0000 | 0.225 | 40.1 | 2346.4 | 2485.3 | 5846.6 | 0 |
| 2 | 41-1 | 0.0006 | 0.310 | 48.7 | 1054.6 | 1154 | 4795 | 0 |
| 3 | 1-19 | 0.0000 | 0.677 | 44.1 | 756 | 697.3 | 4039 | 1 |
| **$M_{SEQ}$**-3:{ 19-9-33-12-50 } | | | | | | | | |
| $N_{Tsk}$ | $W_M$ | Valid | MC | $T_{CPU}$ | $T_M$ | | $T_{Total}$ | |
| 4 | 36 | 1 | 0.220 | 18.9 | 3980 | | 4039 | |
| PP | $E_G$ | $\wp\nabla$ | $\wp C$ | $T_{CPU}$ | $T_\wp$ | $T_\varepsilon$ | $T_{Total}$ | $FM_{re}$ |
| 1 | 19-9 | 0.0000 | 0.238 | 39.3 | 754.6 | 684.8 | 3284.3 | 1 |
| **$M_{SEQ}$**-4:{ 9-14-33-42-2-50 } | | | | | | | | |
| $N_{Tsk}$ | $W_M$ | Valid | MC | $T_{CPU}$ | $T_M$ | | $T_{Total}$ | |
| 5 | 32 | 1 | 1.080 | 26.4 | 2998 | | 3284.3 | |
| PP | $E_G$ | $\wp\nabla$ | $\wp C$ | $T_{CPU}$ | $T_\wp$ | $T_\varepsilon$ | $T_{Total}$ | $FM_{re}$ |
| 1 | 9-14 | 0.0000 | 0.144 | 44.5 | 1014.5 | 1163 | 2269.8 | 0 |
| 2 | 14-33 | 0.0000 | 0.301 | 38.9 | 490.4 | 334.7 | 1779.4 | 1 |
| **$M_{SEQ}$**-5:{ 33-2-48-50 } | | | | | | | | |
| $N_{Tsk}$ | $W_M$ | Valid | MC | $T_{CPU}$ | $T_M$ | | $T_{Total}$ | |
| 3 | 25 | 1 | 0.770 | 26.1 | 1668 | | 1779.4 | |
| PP | $E_G$ | $\wp\nabla$ | $\wp C$ | $T_{CPU}$ | $T_\wp$ | $T_\varepsilon$ | $T_{Total}$ | $FM_{re}$ |
| 1 | 33-2 | 0.0000 | 0.224 | 40.6 | 751.3 | 686.7 | 1028.1 | 1 |
| **$M_{SEQ}$**-6:{ 2-50 } | | | | | | | | |
| $N_{Tsk}$ | $W_M$ | Valid | MC | $T_{CPU}$ | $T_M$ | | $T_{Total}$ | |
| 1 | 9 | 1 | 1.890 | 23.7 | 830 | | 1028.1 | |
| PP | $E_G$ | $\wp\nabla$ | $\wp C$ | $T_{CPU}$ | $T_\wp$ | $T_\varepsilon$ | $T_{Total}$ | $FM_{re}$ |
| 1 | 2-50 | 0.0000 | 0.228 | 39.2 | 746.6 | 828.6 | 281.4 | done |

| Table.2(h). Experiment-10 | | | | | | | | |
|---|---|---|---|---|---|---|---|---|
| **$M_{SEQ}$**-1:{ 1-7-25-10-15-49-9-29-18-30-50 } | | | | | | | | |
| $N_{Tsk}$ | $W_M$ | Valid | MC | $T_{CPU}$ | $T_M$ | | $T_{Total}$ | |
| 10 | 43 | 1 | 0.480 | 19.8 | 10574 | | 10800 | |
| PP | $E_G$ | $\wp\nabla$ | $\wp C$ | $T_{CPU}$ | $T_\wp$ | $T_\varepsilon$ | $T_{Total}$ | $FM_{re}$ |
| 1 | 1-7 | 0.0001 | 0.263 | 44.3 | 1218.3 | 1500.3 | 9581.7 | 0 |
| 2 | 7-25 | 0.0000 | 0.327 | 43.7 | 762.9 | 688.3 | 8818.8 | 1 |
| **$M_{SEQ}$**-2:{ 25-23-47-45-13-14-12-15-38-50 } | | | | | | | | |
| $N_{Tsk}$ | $W_M$ | Valid | MC | $T_{CPU}$ | $T_M$ | | $T_{Total}$ | |
| 9 | 44 | 1 | 0.350 | 23.4 | 8378 | | 8818.8 | |
| PP | $E_G$ | $\wp\nabla$ | $\wp C$ | $T_{CPU}$ | $T_\wp$ | $T_\varepsilon$ | $T_{Total}$ | $FM_{re}$ |
| 1 | 25-23 | 0.0000 | 0.118 | 39.6 | 394.4 | 517.3 | 8424.3 | 0 |
| 2 | 23-47 | 0.0000 | 0.541 | 40.8 | 474.1 | 386.3 | 7950.1 | 1 |
| **$M_{SEQ}$**-3:{ 47-32-46-35-30-28-50 } | | | | | | | | |
| $N_{Tsk}$ | $W_M$ | Valid | MC | $T_{CPU}$ | $T_M$ | | $T_{Total}$ | |
| 6 | 29 | 1 | 0.330 | 21.9 | 7460 | | 7950.1 | |
| PP | $E_G$ | $\wp\nabla$ | $\wp C$ | $T_{CPU}$ | $T_\wp$ | $T_\varepsilon$ | $T_{Total}$ | $FM_{re}$ |
| 1 | 47-32 | 0.0003 | 0.208 | 36.4 | 2044.6 | 2150 | 5905.6 | 0 |
| 2 | 32-46 | 0.0000 | 0.326 | 39.3 | 1670.2 | 1864.7 | 4235.4 | 0 |
| 3 | 46-35 | 0.0000 | 0.480 | 44.2 | 427.2 | 347 | 3808.1 | 1 |
| **$M_{SEQ}$**-4:{ 35-26-49-15-2-46-50 } | | | | | | | | |
| $N_{Tsk}$ | $W_M$ | Valid | MC | $T_{CPU}$ | $T_M$ | | $T_{Total}$ | |
| 6 | 27 | 1 | 0.280 | 27.3 | 3354 | | 3808.1 | |
| PP | $E_G$ | $\wp\nabla$ | $\wp C$ | $T_{CPU}$ | $T_\wp$ | $T_\varepsilon$ | $T_{Total}$ | $FM_{re}$ |
| 1 | 35-26 | 0.0000 | 0.101 | 48.3 | 339.5 | 486.3 | 3468.6 | 0 |
| 2 | 26-49 | 0.0000 | 0.205 | 39.7 | 688.5 | 823.6 | 2780 | 0 |
| 3 | 49-15 | 0.0000 | 0.430 | 44.6 | 418.1 | 333.7 | 2361.8 | 1 |
| **$M_{SEQ}$**-5:{ 15-48-29-12-50 } | | | | | | | | |
| $N_{Tsk}$ | $W_M$ | Valid | MC | $T_{CPU}$ | $T_M$ | | $T_{Total}$ | |
| 4 | 18 | 1 | 0.970 | 19.6 | 1997 | | 2361.8 | |
| PP | $E_G$ | $\wp\nabla$ | $\wp C$ | $T_{CPU}$ | $T_\wp$ | $T_\varepsilon$ | $T_{Total}$ | $FM_{re}$ |
| 1 | 15-48 | 0.0016 | 0.640 | 45.1 | 508.3 | 484.3 | 1853 | 0 |
| **$M_{SEQ}$**-6:{ 48-12-15-50 } | | | | | | | | |
| $N_{Tsk}$ | $W_M$ | Valid | MC | $T_{CPU}$ | $T_M$ | | $T_{Total}$ | |
| 3 | 17 | 1 | 1.220 | 23.4 | 1802 | | 1853 | |
| PP | $E_G$ | $\wp\nabla$ | $\wp C$ | $T_{CPU}$ | $T_\wp$ | $T_\varepsilon$ | $T_{Total}$ | $FM_{re}$ |
| 1 | 48-12 | 0.0000 | 0.403 | 43.9 | 477.4 | 358.3 | 1376.1 | 1 |
| **$M_{SEQ}$**-7:{ 12-49-50 } | | | | | | | | |
| $N_{Tsk}$ | $W_M$ | Valid | MC | $T_{CPU}$ | $T_M$ | | $T_{Total}$ | |
| 2 | 23 | 1 | 0.560 | 20.6 | 1315 | | 1376.1 | |
| PP | $E_G$ | $\wp\nabla$ | $\wp C$ | $T_{CPU}$ | $T_\wp$ | $T_\varepsilon$ | $T_{Total}$ | $FM_{re}$ |
| 1 | 12-49 | 0.0000 | 0.111 | 37.2 | 372 | 419.3 | 1004 | 1 |
| 2 | 49-50 | 0.0000 | 0.093 | 38.1 | 892 | 894.6 | 111.8 | done |

Table.2 clarifies the progress of the architecture in different stages of a specific mission toward furnishing the addressed objectives, in which the mission starts with operation of the TOM-P for generating the initial mission plan that takes maximum use of the available time. Therewith, the LOP-P starts its process by generating time optimum collision free path between locations of the first pair of waypoints from the sequence of nodes. When the generated local path is traversed and the targeted waypoint is visited, the expended time for traversing the generated path gets compared to the expected time, in which the $T_{\varepsilon}$ for each edge is drawn from the estimated mission time $T_{M}$. Mission re-planning flag gets zero in cases that $T_{\wp}$ is smaller than $T_{\varepsilon}$, which means the current mission plan is still valid and optimum, so the LOP-P can continue its process of next pair of waypoints; however, in the opposite case, the Mission re-planning flag gets one, the traversed edges get eliminated from the network and the TOM-P is recalled to reorganize the task sequence and generate a new route from the current waypoint based on updated available time and terrain. This synchronous process between TOM-P and LOP-P frequently repeated until the AUV reaches the destination that means mission success or available time gets minus value, which means the vehicle runs out of battery. As declared in Table.2 and *Fig*.15, in all of the experiments the $T_{Total}$ got a positive value which means no failure is occurred in this simulation. This issue is remarkable for having a confident and reliable mission as the failure is not acceptable for AUVs due to expensive maintenance in sever underwater environment. The mentioned process is clearly presented in each of the above experiments in Table.2.

An efficient synchronization flows between higher and lower level modules to incorporate dynamic changes of the terrain and rearrange priority of the tasks. The best possible performance for the proposed model is completion of the mission with a minimum positive available time, which means the vehicle took maximum use of the available time and terminated its mission before runs out of battery. The update process of the mission available time that presented in *Fig*.15, shows that maximum use of available time has been taken as the available time in the last recall of the LOP-P remarkably approaches to zero in all experiments, which confirms supreme performance of the proposed novel architecture in mission reliability and excellent time management. Moreover, the results presented in Table.2 shows a very short computational time for both TOM-P and LOP-P in all experiments that confirms real-time performance of this architecture, which makes it remarkably appropriate for real-time application.

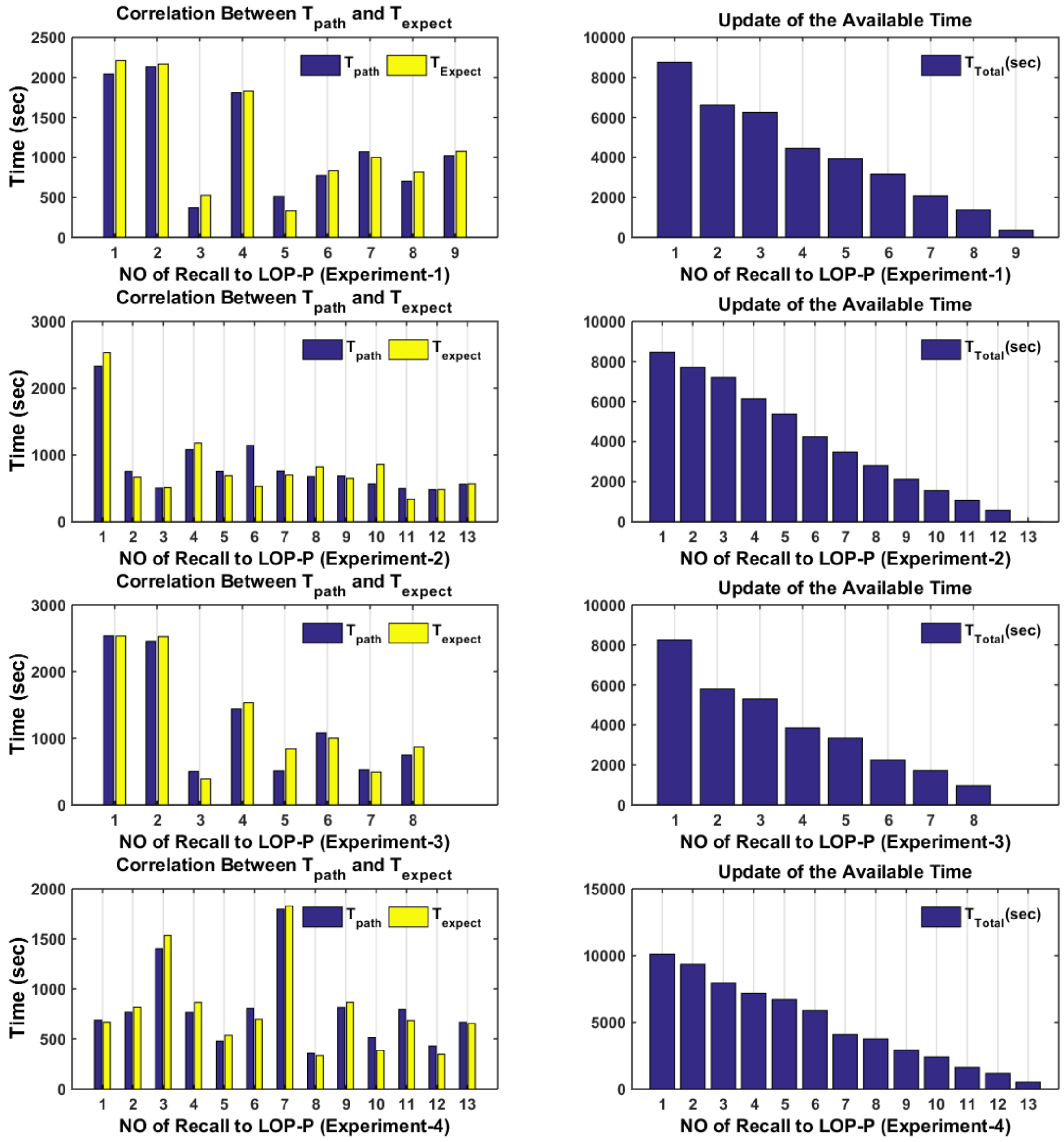

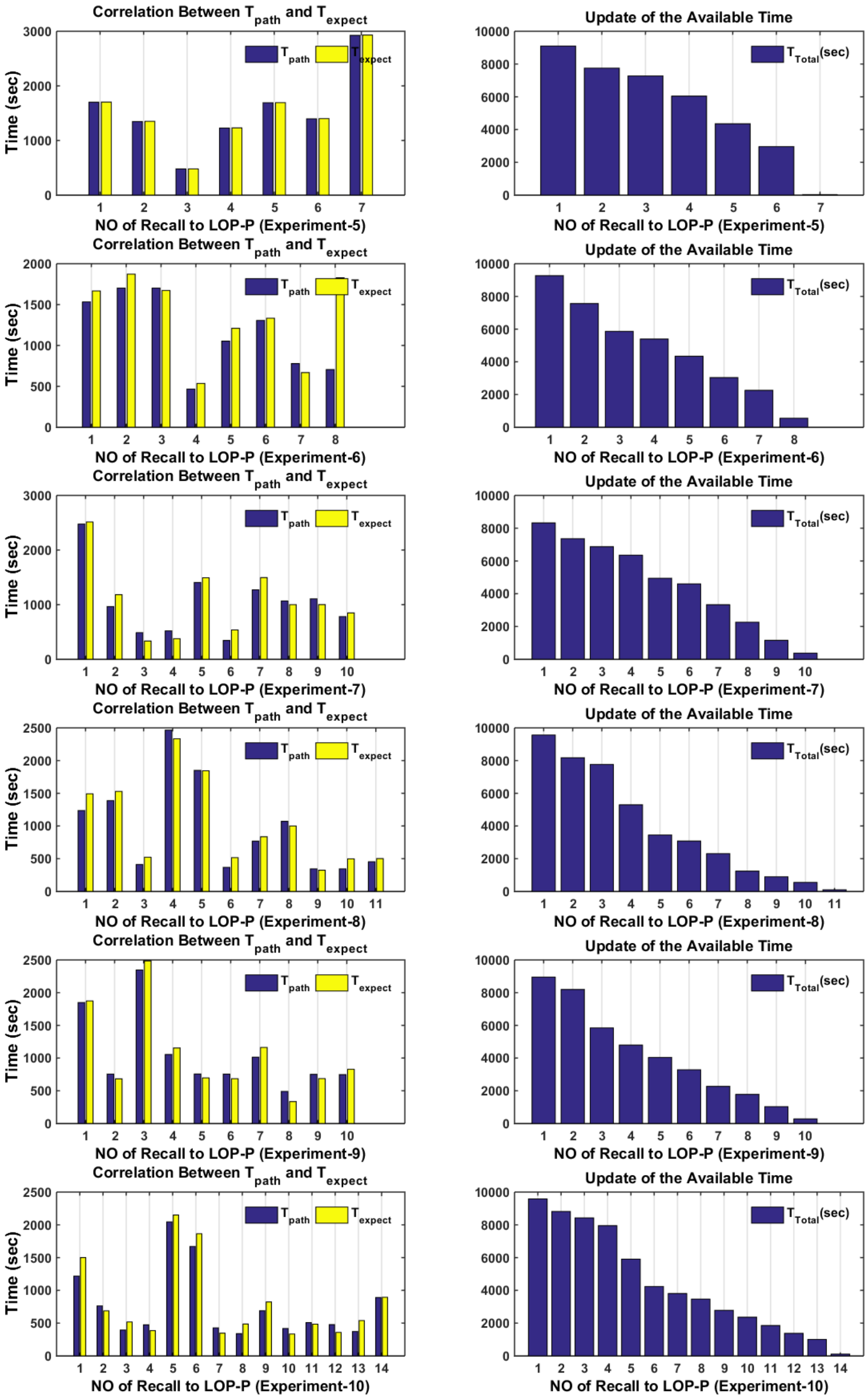

**Fig.15.** Architecture Stability in total time management and balancing correlation between $T_{\wp}$ and $T_{\varepsilon}$ in 10 experiments

As mentioned earlier, re-routing is required any time that $T_{\wp}$ exceeds $T_{\varepsilon}$ and mission available time gets updated after each recall to LOP-P module. It further noted from simulation results in *Fig*.15 that excellent cooperation flows between TOM-P and LOP-P as the correlation between value of $T_{\wp}$ and $T_{\varepsilon}$ is very close and accordant in multiple process of LOP-P in each experiment. Variations of $T_{\wp}$ and $T_{\varepsilon}$ in *Fig*.15 reveals the stability of the architecture as the obtained results are quantitatively very similar in all experiments. This strategy guarantees on time termination of the mission.

## 5 Conclusion and Future Works

Generally, the AUV's control architecture consists of two different execution layers based on its control requirements i.e. deliberative and reactive layers. The deliberative layer manages the concurrent execution of several tasks with different priorities. The reactive layer carries out the real-time reactions to perform quick respond to critical sudden events. These automatic functions are executed in the background and promote AUVs self-management characteristics. To provide a higher level of autonomy, cooperative hybrid architecture has been developed to promote vehicles capabilities in decision-making and situational awareness.

The aim of this research was to introduce a novel synchronous architecture including higher level task-organizer/mission-planner and lower level On-line path planner to provide a comprehensive control on mission time management and performing beneficent mission with the best sequence of tasks fitted to available time, while safe deployment is guaranteed at all stages of the mission. In this respect, the top level module of the architecture (TOM-P) is designed to assign the prioritized tasks in a way that the selected edges in a graph-like terrain lead the vehicle toward its final destination in predefined restricted time. Respectively, in the lower level the LOP-P module handles vehicle safe deployment along the selected edges in presence of severe environmental disturbances, in which the generated path is reactively corrected and refined to cope the sudden changes of the terrain. Precise and concurrent synchronization of the higher and lower level modules is the critical requirement to preserve the accretion and cohesion of the system that promotes stability of the architecture toward furnishing the main objectives addressed above. Developing such a structured modular architecture is advantaged to be upgraded or modified in an easier way that promotes reusability of the system to be implemented on other robotic platforms such as ground or aerial vehicles. In this way, the architecture also can be developed by adding new modules to handle vehicles inner situations, fault tolerant or other consideration toward increasing vehicle's general autonomy. To examine the overall performance of the newly proposed architecture, first the performance of the modules are evaluated separately using ACO algorithm in context of the TOM-P module and FFA algorithm in context of the LOP-P module; afterward, the performance and robustness of the whole system is examined an evaluated through testing of 10 missions' in 10 individual experiments. It is noteworthy to mention from analyse of the results of either TOM-P, LOP-P, and architecture, the proposed strategy enhances the vehicle's autonomy in time management and carrying out safe, reliable and efficient mission for a single vehicle operation in a limited time interval. Besides, it is inferred from analyze of the simulation results that the variation of computational time for operation of both TOM-P and LOP-P modules is settled in a narrow bound in range of seconds for all experiments that confirm accuracy and real-time performance of the system. More importantly, referring to results of 10 experiments in Table.2, the violation value is almost zero for operation of both modules claiming that the system is robust to the variations of the environment while operating in irregularly shaped uncertain spatiotemporal environment in presence of the time-varying water current and other disturbances.

For the future work, it is targeted to integrate the inner modules of the architecture to cover internal situation awareness of the vehicle and promoted vehicles ability in risk management as its failure is not acceptable due to expensive maintenance. Also several other algorithms will be implemented and tested on inner modules of the architecture to investigate the total independency of the architecture's performance from the adopted methods.

## References

[1] Furlong, M.E., Paxton, D., Stevenson, P., Pebody, M., McPhail, S.D., Perrett, J. (2012). Autosub Long Range: A long range deep diving AUV for ocean monitoring. *Autonomous Underwater Vehicles (AUV), 2012 IEEE/OES Conf.*, Southampton, UK, pp. 1–7. (doi:10.1109/AUV.2012.6380737).

[2] Hobson, B., Bellingham, J.G., Kieft, B., McEwen, R., Godin, M., Zhang, Y. (2012). Tethys class long range AUVs-extending the endurance of propeller-driven cruising AUVs from dasys to weeks. *Autonomous Underwater Vehicles (AUV), 2012 IEEE/OES Conference*, pp. 1–8.

[3] Johnson, N., Patron, P., Lane, D. (2007). The importance of trust between operator and auv: Crossing the human/computer language barrier. *IEEE Oceans Europe*, pp. 1159–1164.

[4] Johnson, N.A., Lane, D. (2008). An architecture for planner based control of autonomous vehicles. *27th Workshop of the UK Planning and Scheduling Special Interest Group*, 2008.

[5] Liu, Y., Bucknall, R. (2015). Path planning algorithm for unmanned surface vehicle formations in a practical maritime environment. *Ocean Engineering 97*, pp. 126 –144.

[6] Eichhorn, M. (2015). Optimal routing strategies for autonomous underwater vehicles in time-varying environment. *Robotics and Autonomous Systems*, Vol. 67, pp. 33–43.

[7] Kladis, G.P., Economou, J.T., Knowles, K., Lauber, J., Guerra, T.M. (2011). Energy conservation based fuzzy tracking for unmanned aerial vehicle missions under a priori known wind information. *Engineering Applications of Artificial Intelligence*, Vol. 24 (2), pp. 278–294.

[8] Kwok, K.S., Driessen, B.J., Phillips, C.A., Tovey, C.A. (2002). Analyzing the multiple-target-multiple-agent scenario using optimal assignment algorithms. *Journal of Intelligent and Robotic Systems*, Vol. 35(1), pp. 111– 122.

[9] Liu, L., Shell, D.A. (2012). Large-scale multi-robot task allocation via dynamic partitioning and distribution. *Autonomous Robots*, Vol. 33(3), pp. 291–307.

[10] Higgins, A.J. (2001). A dynamic tabu search for large-scale generalised assignment problems. *Computers & Operations Research*, Vol. 28(10), pp. 1039–1048.

[11] Lysgaard, J., Letchford, A.N., Eglese, R.W. (2004). A new branch-and-cut algorithm for the capacitated vehicle routing problem. *Mathematical Programming*, Vol. 100(2), pp. 423–445.

[12] Al-Hasan, S., Vachtsevanos, G. (2002). Intelligent route planning for fast autonomous vehicles operating in a large natural terrain. *Elsevier Science B.V., Robotics and Autonomous Systems*, Vol.40, pp.1–24.

[13] Gudaitis, M.S. (1994). Multicriteria Mission Route Planning Using a Parallel A* Search. M.Sc. Thesis, Air Force Institute of Technology, *Air University*.

[14] Pereira, A.A., Binney, J., Hollinger, G.A., Sukhatme, G.S. (2013). Risk-aware Path Planning for Autonomous Underwater Vehicles using Predictive Ocean Models. *Journal of Field Robotics*, Vol.30(5), pp.741–762.

[15] Sharma, Y., Saini, S.C., Bhandhari, M. (2012). Comparison of Dijkstra's Shortest Path Algorithm with Genetic Algorithm for Static and Dynamic Routing Network. *International Journal of Electronics and Computer Science Engineering*, ISSN-2277-1956/V1N2, pp.416-425.

[16] M.Zadeh, S., Powers, D., Sammut, K., Lammas, A., Yazdani, A.M. (2015). Optimal Route Planning with Prioritized Task Scheduling for AUV Missions. *IEEE International Symposium on Robotics and Intelligent Sensors*, pp 7-15.

[17] M.Zadeh S, Powers D, Yazdani A.M. (2016). A Novel Efficient Task-Assign Route Planning Method for AUV Guidance in a Dynamic Cluttered Environment. *IEEE Congress on Evolutionary Computation* (*CEC*). Vancouver, Canada. July 24-29.

[18] Carsten, J., Ferguson, D., Stentz, A. (2006). 3D field D*: improved path planning and replanning in three dimensions. *IEEE International Conference on Intelligent Robots and Systems*, pp. 3381-3386.

[19] Koay, T.B., Chitre, M. (2013). Energy-efficient path planning for fully propelled AUVs in congested coastal waters. *Oceans 2013 MTS/IEEE Bergen: The Challenges of the Northern Dimension*.

[20] Petres, C., Pailhas, Y., Evans, J., Petillot, Y., Lane, D. (2005). Underwater path planning using fast marching algorithms. *Oceans Europe Conf*., Brest, France, Vol. 2, pp. 814–819.

[21] Petres, C., Pailhas, Y., Patron, P., Petillot, Y., Evans, J., Lane, D. (2007). Path planning for autonomous underwater vehicles. *IEEE Transactions on Robotics*, Vol.23 (2), pp.331-341.

[22] Roberge, V., Tarbouchi, M., Labonte, G. (2013). Comparison of Parallel Genetic Algorithm and Particle Swarm Optimization for Real-Time UAV Path Planning. *IEEE Transactions on Industrial Informatics*. Vol.9(1), pp.132–141. Doi: 10.1109/TII.2012.2198665.

[23] M.Zadeh S, Powers D, Sammut, K., Yazdani A.M. (2016). Toward Efficient Task Assignment and Motion Planning for Large Scale Underwater Mission. Robotics (cs.RO). arXiv:1604.04854

[24] M.Zadeh, S., Yazdani, A.M., Sammut, K., Powers, D. (2016). AUV Rendezvous Online Path Planning in a Highly Cluttered Undersea Environment Using Evolutionary Algorithms. Robotics (cs.RO). arXiv:1604.07002

[25] M.Zadeh S, Powers D, Sammut, K., Yazdani A.M. (2016). Differential Evolution for Efficient AUV Path Planning in Time Variant Uncertain Underwater Environment. Robotics (cs.RO). arXiv:1604.02523

[26] Nikolos, I.K., Valavanis, K.P., Tsourveloudis, N.C., Kostaras, A.N. (2003). Evolutionary Algorithm Based Offline/Online Path Planner for UAV Navigation. *IEEE Transactions on Systems, Man, and Cybernetics, Part B*, Vol.33(6), pp.898–912. Doi: 10.1109/tsmcb.2002.804370.

[27] Alvarez, A., Caiti, A., Onken, and R. (2004). Evolutionary path planning for autonomous underwater vehicles in a variable ocean. *IEEE Journal of Oceanic Engineering*, Vol. 29(2), pp.418–429. doi:10.1109/joe.2004.827837.

[28] Ataei, M., Yousefi-Koma, A. (2015). Three-dimensional optimal path planning for waypoint guidance of an autonomous underwater vehicle. *Robotics and Autonomous Systems*, Vol. 67, pp 23–32.

[29] Zeng, Z., Sammut, K., Lammas, A., He, F., Tang, Y. (2014). Shell space decomposition based path planning for AUVs operating in a variable environment. *Ocean Engineering*, Vol. 91, pp. 181-195.

[30] M.Zadeh S, Powers D, Sammut, K., Yazdani A.M. (2016). A Novel Versatile Architecture for Autonomous Underwater Vehicle's Motion Planning and Task Assignment. Robotics (cs.RO). arXiv:1604.03308v2

[31] M.Zadeh S, Powers D, Sammut, K., Yazdani A.M. (2016). Biogeography-Based Combinatorial Strategy for Efficient AUV Motion Planning and Task-Time Management. Robotics (cs.RO). arXiv:1604.04851

[32] M.Zadeh S, Sammut, K, Powers DMW, Yazdani A.M. (2016). 2016. A Hierarchal Planning Framework for AUV Mission Management in a Spatio-Temporal Varying Ocean. Robotics (cs.RO). arXiv: 1604.07898

[33] http://www.bom.gov.au/australia/satellite/

[34] Smith, R.N., Chao, Y., Li, P.P., Caron, D.A., Jones, B.H., Sukhatme, G.S. (2010). Planning and implementing trajectories for autonomous underwater vehicles to track evolving ocean processes based on predictions from a regional ocean model. *International Journal of Robotics Research*, Vol.29(12), pp.1475–1497.

[35] Oke, P.R., Brassington, G.B., Griffin, D.A., Schiller, A. (2008). The Blue link ocean data assimilation system (BODAS). *Journal of Ocean Modelling*, Vol. 21(1–2), pp.46–70.

[36] Fossen, T.I., (2002). Marine Control Systems: Guidance, Navigation and Control of Ships, Rigs and Underwater Vehicles. Marine Cybernetics Trondheim, Norway.

[37] Dorigo, M., Di Caro, G., Gambardella, L.M., (1999). Ant algorithms for discrete optimization. *Journal of Artificial Life*, Vol.5(2), pp.137 – 172.

[38] Dong, H., Zhao, X., Qu, L., Chi, X., Cui. X. (2014) Multi-Hop Routing Optimization Method Based on Improved Ant Algorithm for Vehicle to Roadside Network. *Journal of Bionic Engineering,* Vol.11, pp. 490-496. Online publication date: 1-Jul-2014.

[39] Euchi, J., Yassine, A., Chabchoub. H. (2015). The dynamic vehicle routing problem: Solution with hybrid metaheuristic approach. *Swarm and Evolutionary Computation*. Vol.21, pp.41-53.

[40] Pendharkar, P.C. (2015). An ant colony optimization heuristic for constrained task allocation problem. *Journal of Computational Science,* Vol.7, pp.37-47.

[41] Yang, X.S. (2009). Firefly algorithms for multimodal optimization. *5th Symposium on Stochast*ic Algorithms, Foundations and Applications, (Eds. O. Watanabe and T. Zeugmann), Lecture Notes in Computer Science, 5792: 169-178 (2009).

[42] Yang, X.S. (2010). Nature-Inspired Metaheuristic Algorithms. *Luniver Press*, UK. 2nd edition, 2010.

[43] M.Zadeh S, Powers D, Sammut, K., Yazdani A.M. (2016). An Efficient Hybrid Route-Path Planning Model for Dynamic Task Allocation and Safe Maneuvering of an Underwater Vehicle in a Realistic Environment. Robotics (cs.RO). arXiv:1604.07545

[44] Yang, X.S., He, X. (2013). Firefly Algorithm: Recent Advances and Applications. *International Journal of Swarm Intelligence*, Vol. 1(1), pp. 36-50.